\documentclass[letterpaper]{article} 
\usepackage[submission]{aaai2026}  
\usepackage{times}  
\usepackage{helvet}  
\usepackage{courier}  
\usepackage[hyphens]{url}  
\usepackage{graphicx} 

\usepackage{amsmath}
\usepackage{amssymb} 

\usepackage{kotex}
\usepackage{natbib}  
\usepackage{caption} 
\usepackage{algorithm}
\usepackage{algorithmic}

\usepackage{enumitem}
\usepackage{multicol}

\usepackage{lipsum}
\usepackage{amsmath}  
\usepackage{amssymb}        
\usepackage{amsfonts}      
\usepackage{verbatim}      
\usepackage{newfloat}
\usepackage{listings}
\DeclareCaptionStyle{ruled}{labelfont=normalfont,labelsep=colon,strut=off} 
\floatstyle{ruled}
\newfloat{listing}{tb}{lst}{}
\floatname{listing}{Listing}
\usepackage{booktabs}

\title{Pay What LLM Wants:\\\protect Can LLM Simulate Economics Experiment with 522 Real-human Persona?}

\affiliations{
    \textsuperscript{\rm 1}Department of Artificial Intelligence, Chung-Ang University\\
    84 Heukseok-ro, Dongjak-gu, Seoul 06912, Republic of Korea\\
    \{chlwnsgur129,phchu0429,hyebeens819,bgnkim\}@cau.ac.kr\\[4pt]   

    \textsuperscript{\rm 2}Department of Arts and Cultural Management, Chung-Ang University\\
    84 Heukseok-ro, Dongjak-gu, Seoul 06912, Republic of Korea\\
    wplhm@cau.ac.kr
    
    \textsuperscript{\rm 3}School of Economics, Chung-Ang University\\
    84 Heukseok-ro, Dongjak-gu, Seoul 06912, Republic of Korea\\
    hyunjjin@cau.ac.kr
}

\author{
    Junhyuk Choi\textsuperscript{\rm 1$^\dagger$},
    Hyeonchu Park\textsuperscript{\rm 1},
    Haemin Lee\textsuperscript{\rm 2}
    Hyebeen Shin\textsuperscript{\rm 1},
    Hyun Joung Jin\textsuperscript{\rm 3},
    Bugeun Kim\textsuperscript{\rm 1$^\ast$}
}

\usepackage{bibentry}

\begin{document}

\maketitle

\begin{abstract}
Recent advances in Large Language Models (LLMs) have generated significant interest in their capacity to simulate human-like behaviors, yet most studies rely on fictional personas rather than actual human data. We address this limitation by evaluating LLMs' ability to predict individual economic decision-making using Pay-What-You-Want (PWYW) pricing experiments with real 522 human personas. Our study systematically compares three state-of-the-art multimodal LLMs using detailed persona information from 522 Korean participants in cultural consumption scenarios. We investigate whether LLMs can accurately replicate individual human choices and how persona injection methods affect prediction performance. Results reveal that while LLMs struggle with precise individual-level predictions, they demonstrate reasonable group-level behavioral tendencies. Also, we found that commonly adopted prompting techniques are not much better than naive prompting methods; reconstruction of personal narrative nor retrieval augmented generation have no significant gain against simple prompting method. We believe that these findings can provide the first comprehensive evaluation of LLMs' capabilities on simulating economic behavior using real human data, offering empirical guidance for persona-based simulation in computational social science.
\end{abstract}

\section{Introduction}
Recent advances in Large Language Models (LLMs) have sparked growing interest in their ability to reproduce social and psychological behaviors, leading to active research on simulating human-like judgment and choice patterns \citep{park2024generative,park2023generative,wang2023rolellm,cheng2023compost,huang2024humanity,choi2025examiningidentitydriftconversations}. Of particular note is the approach of injecting personas into LLMs to mimic human attitudes, personality traits, value judgments, and choice behaviors. However, most prior studies have focused on evaluating general tendencies without precise behavioral comparisons with actual humans, employing methods such as generating fictional personas to simulate human social phenomena, creating personas using demographic information, or extracting personas from existing QA datasets \citep{horton2023large,hewitt2024predicting,leng2023llm,ross2024llm,prospect,xie2024can}.


To our knowledge, \citet{park2024generative} is the only study analyzing actual human behavior, achieving 85\% accuracy in reproducing General Social Survey responses from interview data. However, this study focused on psychological surveys rather than economic decision-making involving monetary costs. Simple attitudinal survey can elicit idealistic responses without realistic constraints, making it difficult to reflect actual decision making. Economic decision-making requiring willingness to pay involves complex cognitive processes where budget constraints, opportunity costs, and cultural interact, providing a more precise foundation for evaluating LLMs' human-like behavior reproduction.

To bridge this gap, we designed economic decision-making experiment using Pay-What-You-Want (\citet{kim2009pay}; PWYW) pricing, where consumers voluntarily determine payment amounts. This involves complex psychological factors including willingness to pay, income, cultural values, and individual value systems \citep{kim2009pay,gerpott2017pay}, providing a valid framework for analyzing individual decision-making and evaluating how sophisticatedly LLMs can simulate actual individual human choices. Unlike previous studies evaluating general tendencies with fictional personas, we apply identical experimental structures to LLMs using detailed persona information from 522 actual participants and their decision-making data, analyzing how persona structure affects LLM choices and evaluating both group-level choice tendencies and individual-level prediction consistency. Therefore, this study evaluates how precisely LLMs can predict individual behaviors based on actual human personas within PWYW contexts. We systematically examine LLMs' human-like behavior reproduction using various models and persona injection methods through two key research questions.

\begin{itemize}
    \item \textbf{RQ1.} Can LLMs predict economic decision of human individual in sequential or human-guided condition?
\end{itemize}

The first research question examines whether LLMs can reproduce decision-making patterns similar to corresponding human participants when injected with individual persona information. We construct individualized personas encompassing each participant's educational background, cultural experiences, value systems, and consumption tendencies, then input these into various LLMs to predict willingness-to-pay amounts and choice behaviors for the same cultural products. Based on prior studies \citep{lampinen2022can, wei2025shadows} showing that errors may accumulate as LLMs construct subsequent judgments based on previous responses, we analyze the impact of response accumulation on prediction performance. We establish two experimental conditions: 1) Sequential Condition, where LLMs maintain their previous responses and answer all items sequentially, and 2) Human-guided Condition, where actual human responses are inserted into the conversation history to inform subsequent questions. Through this design, we directly compare LLMs' predictions with actual human responses and quantitatively evaluate both individual-level accuracy and group-level tendencies.

\begin{itemize}
    \item \textbf{RQ2.} How do persona formats and prompting methods affect LLMs' ability in simulating human behavior?
\end{itemize}

The second research question analyzes how LLMs' prediction performance varies depending on persona composition and injection methods. While previous studies show LLMs respond sensitively to contextual changes \citep{sclar2023quantifying,zhuo2024prosa,razavi2025benchmarking}, no systematic comparison exists regarding which methods are more effective for predicting human behavior. This study establishes persona injection along two axes: (1) Format: Survey vs. Storytelling (i.e., biographical narratives), and (2) Prompting Methods: base prompt, CoT \citep{wei2022chain}, RAG \citep{lewis2020retrieval}, and Few-shot Prompting \citep{brown2020language}. 
By applying identical conditions across different models, we analyze performance differences between models and evaluate how method variations are reflected in individual-level accuracy and group-level tendencies.

Based on this research design, this study provides two major contributions. First, we systematically validate the extent to which LLMs can precisely mimic individual human decision-making at both group and individual levels, based on sophisticated persona and decision-making data collected from actual human participants. This represents one of the first studies to evaluate LLMs' ability to predict specific humans' actual choice behaviors, advancing the precision of LLM human simulation research to the next level. Second, by comparing how LLMs' behavioral prediction performance varies according to persona composition and injection methods even with identical information, we provide empirical evidence for appropriate LLM utilization methods for future human-like behavior generation and simulation.


\section{Related Work}
Recent research has increasingly focused on evaluating how precisely LLMs can replicate human-like social and psychological behaviors. We categorize these studies into three types: persona-free attempts, attempts with simulated personas, and those with real personas.


\paragraph{Persona-free Investigation with Social Psychology:}
To quantitatively assess LLM social behaviors, studies have adopted established social game paradigms including the Public Goods Game, Ultimatum Game, and organizational collaboration scenarios to evaluate cooperative tendencies, altruism, and strategic capabilities \citep{Simulating_Cooperative, Cooperative_Prosocial}. These approaches assign specific roles to models or structure cooperative frameworks through prompts, simulating human-like teamwork and decision-making patterns. While significant in demonstrating that LLMs can modulate behavior based on contextual conditions, most analyses focus on aggregate group trends, showing multi-agent systems approximate human cooperative distributions better than single models. However, findings remain limited to group-level statistical similarities, leaving individual persona-level predictive precision largely unexamined.



\paragraph{Evaluation Attempts with Simulated Personas:}
Based on the information type provided to the model, these personas fall into three categories: personality-based, demographic-based, and narrative-based. Some researchers attempted to input personality-based personas from social psychology as prompts to induce distinct behavioral patterns \citep{zhang-etal-2024-exploring, Emergent_social}. For example, \citeauthor{zhang-etal-2024-exploring} examined how agents with different personality traits engage in negotiation and debate. 
%
%
Other researchers attempted to design demographic-based personas for simulating variation in attitudes toward social issues \citep{AgentSociety, Moral_Machine}. For example, \citeauthor{AgentSociety} injected demographic profiles into over 10,000 LLM agents to analyze decision-making distributions in contexts such as basic income policies and disaster response. 
%
Moreover, another group of researchers attempted to utilize narrative-based personas including specific memories, goals, and daily routines to simulate long-term and personalized interactions \citep{park2023generative, Can_LLMs_Trust}. \citeauthor{park2023generative} provided LLM agents with background stories and tracked their autonomous behaviors within a virtual town, demonstrating the potential for sustained, agent-specific behavioral patterns. 
%
%
%
While these approaches have been effective in diversifying LLM behavior through persona injection, they remain limited by the synthetic nature of the input. That is, the personas are not grounded in real human data, and thus miss achieving a one-to-one correspondence with actual human attitudes and contextual experience.
\begin{figure*}[htbp]  
    \centering
    \includegraphics[width=0.9\textwidth]{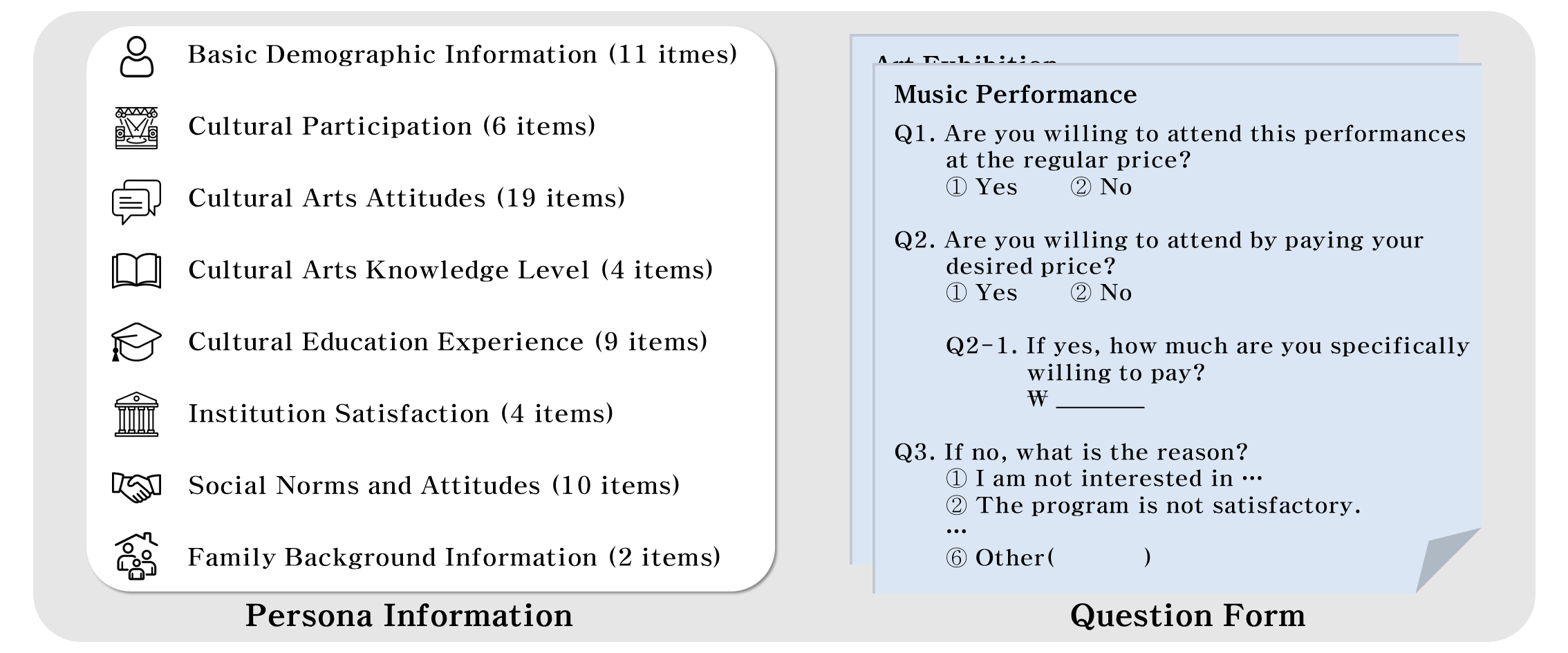}  
    \caption{Overview of persona information categories and experimental question form.}  
    \label{fig:form_fig}  
\end{figure*}


\paragraph{Evaluation Attempts with Real Personas:}
While simulated personas have effectively induced behavioral variation in LLMs, they offer limited precision in predicting individual behavior. More recently, \citet{park2024generative} evaluated LLM-human alignment using real persona data from interviews, comparing outputs across the General Social Survey, Big Five traits, and simple economic games. However, this work focused primarily on attitudinal surveys, not high-stakes economic decisions involving monetary consequences—contexts requiring higher-order reasoning shaped by budget constraints, opportunity costs, and cultural norms. Furthermore, it did not examine how different persona injection methods impact predictive accuracy.

Though previous studies identified valuable insights using different personas, no existing studies have systematically evaluated LLMs' ability to predict realistic economic consumption choices. In addition, they have less analyzed how variations in persona construction and injection strategies influence behavioral alignment. To address these gaps, investigating whether LLMs can simulate economic human behavior with real-world persona is required. To conduct such a simulation, we adopt a human experiment about willingness-to-pay, as illustrated in the next section.

\section{Human Experiment} 
This section describes the human experiment that provides the baseline data for evaluating LLMs' individual behavioral prediction capabilities. The experiment is based on data collected from \citet{anonymous}. In this section, we briefly introduce (1) the overall design of conducted PWYW experiment, (2) the collected persona items and components, and (3) the questions for inspecting economic decision-making.

\subsection{Pay What You Want}
This study adopted Pay-What-You-Want (PWYW) method to measure participants' economic choices and willingness to pay. PWYW is a participatory pricing mechanism where consumers voluntarily determine payment amounts based on their perceived value rather than fixed prices, and has been widely used as a representative experimental tool for behaviorally measuring willingness to pay in economics and consumer behavior research \citep{kim2009pay,gerpott2017pay}. This method goes beyond simple product selection, requiring consumers to comprehensively consider various internal factors such as their preferences, economic constraints, fairness perceptions, social norms, and value systems when determining payment amounts. PWYW effectively implements experimental decision-making situations where these complex factors interact to influence willingness to pay. Particularly, by applying the PWYW context to cultural product consumption situations, \citet{anonymous} designed the experiment to observe how participants' cultural tastes, experiences, knowledge, educational levels, and so-called cultural capital are reflected in actual consumption decisions. This enables collection of more realistic and coherent economic choice data rather than simple preference responses, providing experimental conditions suitable for evaluating LLMs' ability on simulating human behavior.

\subsection{Persona Information}
\citet{anonymous} collected personas from 522 Korean adults (Male 254, Female 268; Ages between 30-39). As shown in Figure \ref{fig:form_fig}, they investigated a total of 65 items including: basic demographic information, attitudes and preferences toward cultural arts, cultural arts knowledge level, cultural education experience, institutional usage experience and satisfaction, social norms and attitudes, and family background information. Through this comprehensive approach, they constructed personas that holistically reflect each individual's cultural capital level and socioeconomic background. We asked the authors and received the data. For detailed persona questionnaire, see supplementary material.

\subsection{Question Form}
To construct realistic PWYW decision-making situations for cultural products, this study designed two scenarios: art exhibitions and music performances. Each scenario was structured to precisely measure participants' willingness to pay and choice behaviors, as presented in Figure~\ref{fig:form_fig}. Specifically, three key questions were used to elicit participants' economic decision-making in each scenario. First, a binary question asking whether they are willing to attend the performance at the regular price. Second, a binary question asking whether they are willing to attend by paying their desired price. Third, for those who expressed willingness to attend in the second question, an open-ended numerical question asking how much they are willing to pay specifically. Additionally, multiple-choice questions were presented to identify reasons for unwillingness to attend, enabling analysis of how accurately LLMs can predict individual responses across various question formats. For the original korean questionnaire, refer to Supplementary Material.

\subsection{Statistical Tests Conducted in the Experiment}
Group-level human tendencies were estimated using the Heckman two-step model \citep{heckman1976common}. In the first step, a probit regression was conducted with the second yes/no items for each scenario (\texttt{Q2}) as dependent variables and persona attributes as independent variables to estimate the probability of viewing intention. In the second step, a linear regression was performed using the willingness-to-pay amounts (\texttt{Q2.1}) as dependent variables, limited to participants agreed in \texttt{Q2}. The inverse Mills ratio calculated from the first stage was incorporated to correct for self-selection bias. The model accounts for potential correlation between error terms in both stages, and sample selection bias occurs when this correlation is non-zero. For detailed definitions, see supplementary material.

\section{RQ1 : Sequential vs Human-guided}
Using the above experiment, this section investigates whether LLMs can accurately predict individual human decisions under Sequential versus Human-guided interaction conditions. We address the detailed experimental procedures for validating this question, the characteristics of models used for evaluation, analytical methods, and results of each model's prediction performance against human responses.

\subsection{Experiment Setup}
We evaluated LLMs' human simulation performance by completely replicating the structure of the human experiment. Each LLM was provided with real persona information of actual human participants, and all elements of the questionnaire used in the human experiment were presented identically. Specifically, we provided all contextual information identical to humans, including survey introductions, detailed descriptions of exhibitions and performances, images provided to human participants, and hypothetical situation settings. Given that all participants in this study were Korean, and considering prior research \citep{leng2024can, verma2023large} showing that linguistic variations can affect results, we used all survey items, system prompts, and questions input to LLMs in their original Korean form. The specific experimental procedure involved first presenting each participant's persona information to the LLM as a System Prompt, appending the instruction `Think of yourself as a person of a given persona and answer' in Korean, and then asking questions in the same order as provided to humans.

To analyze the impact of previous responses on subsequent answers, we established two experimental conditions. The first was the \textit{Sequential Condition}, where previous responses generated by the LLM were maintained as-is, allowing all questions to proceed sequentially with LLM responses accumulating. The second was the \textit{Human-guided Condition}, where we inserted the actual responses of the corresponding human participants into the conversation log from the second question onward to replace the LLM's previous responses. This ensured that the LLM's response to each question was based on actual human responses rather than its own previous answers, enabling measurement of more independent and accurate prediction performance.


\subsection{Evaluation}
We evaluated whether LLMs can predict human behavior with two aspects: \textit{individual-level accuracy} and \textit{group-level tendencies}. First, individual-level accuracy measures how accurately LLMs predicted individual participants' choices. Typically, the accuracy is measured with the proportion of cases where human responses and LLM responses completely match. Specifically, we calculate: (1) accuracy rate for each individual item, (2) accuracy rate for all binary items (\texttt{Q1} and \texttt{Q2}), and (3) Overall accuracy for correctly answering all three items, providing a multifaceted evaluation of LLM prediction performance. 

Second, group-level tendencies measures the overall similarity between human data and simulated data by LLMs. Typically, we compared patterns of two regression models in estimating the influence of persona variables. Three metrics are used for this purpose: (1) \textit{Coefficient Sign Agreement} (CSA) is the proportion of regression coefficients whose signs are matched, showing how similarly two models judged the direction of variable influence. (2) \textit{Statistical Significance Agreement} (SSA) is the proportion of variables that are statistically significant in both models or insignificant in both, indicating the degree of agreement in important variable selection. (3) \textit{Jaccard Index} \citep{jaccard1908nouvelles} calculates the overlap between two models using the set of significant variables, expressing how much the important variable sets themselves overlap as a value.


\subsection{Models}
In this experiment, we selected three state-of-the-art multimodal LLMs capable of simultaneously processing text and images as evaluation targets: GPT-4o \citep{hurst2024gpt}, Llama-3.2-90B-Vision-Instruct \citep{dubey2024llama}, and Qwen2.5-VL-72B-Instruct \citep{bai2025qwen2}. All models are based on multimodal architectures that can integratively process images and text, and also demonstrate good performance in processing Korean, the experimental language of this study. Particularly, on a popular Korean benchmark, Hae-Rae benchmark \citep{son2023hae}, these models achieved good scores: GPT-4o (0.836), Llama-3.2 (0.738), and Qwen2.5 (0.636), respectively. GPT-4o was accessed through the official OpenAI API, while the other two models were accessed through the OpenRouter API. To ensure response consistency, the temperature was fixed at 0 for all.


\subsection{Results and Discussion}
In this section, we analyzes the experimental results regarding two conditions and inter-model comparisons, discussing the similarity and agreement between humans and LLMs.

\begin{table}[t]
\centering
\footnotesize
\label{tab:rq1_full}
\begin{tabular}{l|c@{\;\;}c@{\;\;}c|c@{\;\;}c@{\;\;}c}
\toprule
 & \multicolumn{3}{c|}{\textbf{Sequential}} 
 & \multicolumn{3}{c}{\textbf{Human-guided}} \\

\textbf{Metric} 
& \textbf{GPT} & \textbf{Llama} & \textbf{Qwen}
& \textbf{GPT} & \textbf{Llama} & \textbf{Qwen} \\
\midrule
\multicolumn{7}{l}{\textbf{A. Accuracy by Question (Individual-level)}} \\
M\_Q1               &  62.26   &  60.15   &  61.11   &  62.07 & 63.60 & 62.26 \\
M\_Q2               &  57.28   &  78.54   &  75.10   &  70.50 & 80.84 & 76.82 \\
M\_Q2.1          &  5.43   &  22.25   &  9.78   &  15.65 & 26.41 & 33.74 \\
M\_Q3    &   11.36 & 0.00 & 0.00 & 31.86 & 9.73 & 29.20 \\
TF Total      & 44.06 & 53.83 & 46.17 & 45.98 & 55.36 & 45.79 \\
M\_total       &  6.90   &  11.49   &  4.79   &   6.70 & 11.11 & 15.13 \\
\midrule
A\_Q1               &  64.75   &  37.55   &  66.28   &  65.13 & 55.17 & 65.71 \\
A\_Q2               &  67.82   &  62.84   &  64.75   &  76.44 & 73.37 & 65.71 \\
A\_Q2.1          &  16.23   &  9.6   &  4.28   &  28.44 & 23.24 & 24.16 \\
A\_Q3               & 16.73 & 0.19 & 0.13  & 0.00 & 35.38 & 36.41 \\
TF Total         & 45.59 & 32.95 & 46.55 & 44.44 & 42.72 & 40.04 \\
A\_total       &  12.45   &  5.94   &  1.72   &  17.05 & 16.09 & 13.79 \\
\midrule
\textbf{Total Acc} &  4.60   &  0.77   &  0.77   &   4.02 &  0.96 &  4.41 \\
\midrule
\multicolumn{7}{l}{\textbf{B. Coefficient Similarity to Human (Group-level)}} \\
CSA      
&  0.62   &  0.55  & 0.66   &  0.72   &  0.86   & 0.76 \\
SSA
&  0.86   & 0.83 &  0.72  &  0.93   &  0.90   & 0.86 \\
Jaccard  &  0.43   & 0.00 & 0.27   &  0.67   &  0.57   & 0.20 \\
\bottomrule
\end{tabular}
\caption{LLM performance across experimental conditions for music performance (M) and art (A) domains.}
\label{tab:rq1_result}
\end{table}

\begin{figure}[t]  
    \centering
    \includegraphics[width=0.42\textwidth]{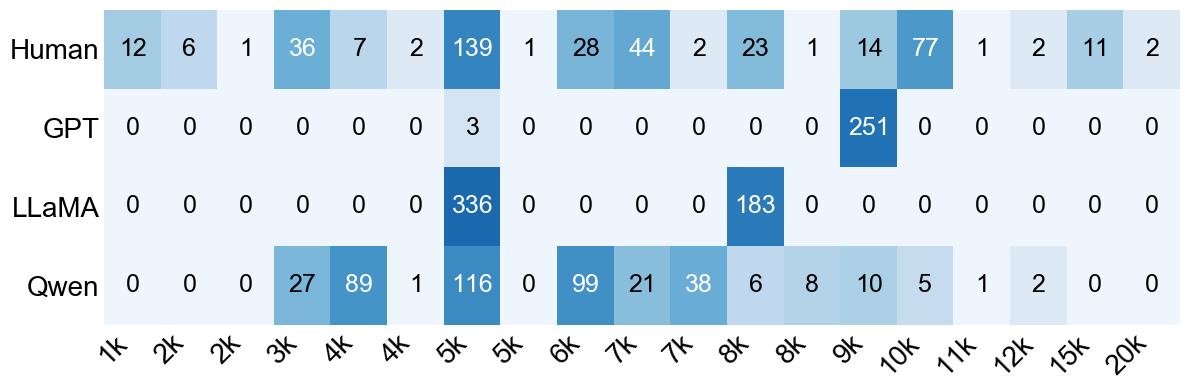}  
    \caption{Response distribution in Sequential Condition}  
    \label{fig:form1}  

    \includegraphics[width=0.42\textwidth]{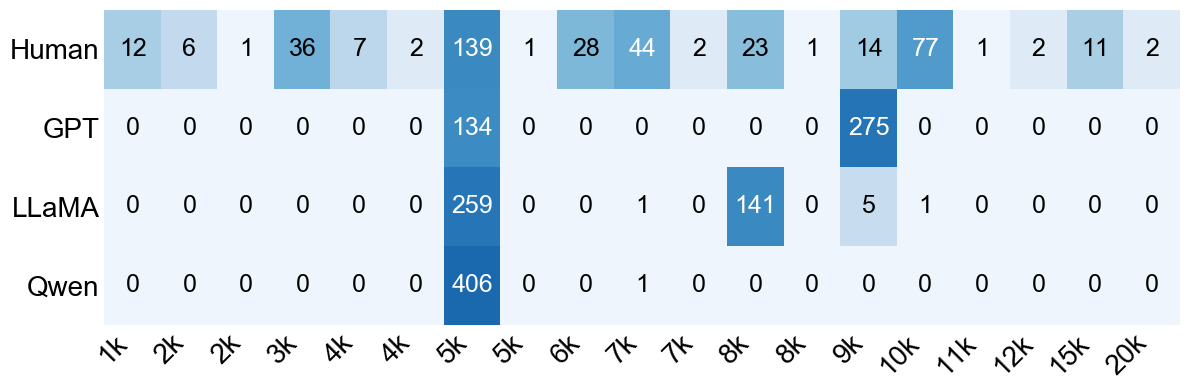}  
    \caption{Response distribution in Human-Guided condition}  
    \label{fig:form2}  
\end{figure}

\paragraph{Effect of Interaction Condition:}
Comparing the performance differences between the two interaction conditions, the Human-guided Condition showed overall superior prediction performance compared to the Sequential Condition. This was somewhat expected, as the Human-guided condition is designed to allow LLMs to reference actual human response flows for each question, enabling reasoning to proceed in a manner closer to human judgment pathways. These results suggest that providing actual human responses as examples may produce effects similar to few-shot prompting, providing a foundation for exploring the effectiveness of various prompting techniques in RQ2. 

As shown in Table \ref{tab:rq1_result}, the Human-guided condition demonstrated higher accuracy than the Sequential condition in terms of individual-level accuracy. The difference between conditions was particularly pronounced in subjective or open-ended questions compared to binary-choice items. For example, in Qwen's case, the accuracy for subjective questions (\texttt{Q2.1}) was only 9.78\% and 4.28\% in the Sequential condition, but improved significantly to 33.74\% and 24.16\% in the Human-guided condition. This suggests that when LLMs rely on their own generated response flows (Sequential Condition), errors may accumulate or reasoning may proceed in directions that diverge from human response patterns. However, we can confirm that the Overall Accuracy was generally very low across all models. Meanwhile, analysis of group-level tendencies showed that both conditions performed well overall, with the Human-guided condition consistently showing higher prediction accuracy compared to the Sequential condition. The Human-guided condition scored higher across all major indicators including CSA, SSA, and Jaccard index.

These results provide two major findings. First, in Human-guided condition, actual human responses were provided as context for each item, enabling LLMs to form more coherent reasoning pathways, which led to higher prediction accuracy at both group and individual levels. This demonstrates that providing actual response examples is effective for improving LLMs' behavioral prediction accuracy, suggesting the validity of example-based approaches such as few-shot. Second, while LLMs still showed difficulties in precisely replicating individual human choices, they demonstrated similar tendencies to humans at the group level.


\paragraph{Comparison Models:}
Based on the results from Table \ref{tab:rq1_result}, Figures \ref{fig:form1} and \ref{fig:form2} when comparing accuracy across models, GPT-4o showed consistently high performance in both item-level accuracy (\texttt{M\_Q1} 62.07\%, \texttt{M\_Q2} 70.50\%, \texttt{A\_Q1} 65.13\%, \texttt{A\_Q2} 76.44\%) and group-level similarity indicators (CSA 0.72, SSA 0.93, Jaccard 0.67) in the Human-guided condition, demonstrating the most human-like patterns in both choice outcomes and variable structures. LLaMA achieved the highest variable direction agreement (0.86) in the Human-guided condition. But overall accuracy was only 0.96\%. This suggests that while the model excels at identifying the direction of variable influence, it lacks consistency in translating this understanding into individual choice predictions. Qwen showed similar Overall Accuracy to GPT-4o at 4.41\% in the Human-guided condition, but had a low Jaccard score of 0.20, indicating that its important variable set differed significantly from humans.

Figures \ref{fig:form1} and \ref{fig:form2} reveal particularly notable model differences in response distribution in terms of price (\texttt{Q2.1}). Figure \ref{fig:form1} shows that human responses have high diversity, while LLM responses tend to cluster at specific price points. In Figure \ref{fig:form2}, this clustering pattern becomes even more pronounced, with LLMs showing much more concentrated response distributions compared to the diverse human response patterns. For instance, in Sequential condition, Figure \ref{fig:form1} show that Qwen provided response distributions with some variance similar to humans, but when switched to Human-guided condition, the response distribution became much more concentrated at specific price points. Nevertheless, the accuracy for the same questions increased significantly, as shown in the transition from Figure \ref{fig:form1} to \ref{fig:form2}. This demonstrates that actual human responses had a positive impact on the model's prediction performance, though at the cost of response diversity.


\begin{table*}[t]
\centering
\footnotesize
\label{tab:rq1_full}
\begin{tabular}{l|c@{\;\;}c@{\;\;}c|c@{\;\;}c@{\;\;}c|c@{\;\;}c@{\;\;}c|c@{\;\;}c@{\;\;}c}
\toprule
 
 & \multicolumn{3}{c|}{\textbf{Basic Prompt}} 
 & \multicolumn{3}{c}{\textbf{CoT Prompt}}& \multicolumn{3}{c}{\textbf{RAG Prompt}} & \multicolumn{3}{c}{\textbf{Few-Shot}} \\

\textbf{Metric} 
& \textbf{GPT} & \textbf{Llama} & \textbf{Qwen}
& \textbf{GPT} & \textbf{Llama} & \textbf{Qwen} & \textbf{GPT} & \textbf{Llama} & \textbf{Qwen} & \textbf{GPT} & \textbf{Llama} & \textbf{Qwen} \\
\midrule
\multicolumn{13}{l}{\textbf{Survey Format}} \\
\multicolumn{7}{l}{\textbf{A. Accuracy by Question (Individual-level)}} \\
M\_Q1               &     62.07 & 63.60 & 62.26  & 65.13  & 63.03 & 65.71 & 62.07 & 60.15 & 62.26 & 66.09 & 60.15 & 64.56\\
M\_Q2               &     70.50 & 80.84 & 76.82  & 74.90  & 78.35 & 78.74 & 72.41 & 80.27 & 71.65 & 70.69 & 78.74&73.37\\
M\_Q2\_1          &    15.65 & 26.41 & 33.74  & 21.27  & 12.96  & 22.74 & 15.40 & 31.78 & 33.99 & 20.05&6.85&20.05\\
M\_Q3    &    31.86 & 9.73 & 29.20  &  31.86  & 23.01 & 28.32 & 31.86 & 28.32 & 27.43 & 32.74&14.16&24.78\\
TF Total     &    45.98 & 55.36 & 45.7  & 55.56 & 53.83 & 51.34 & 47.32 & 47.51 & 48.28 & 49.43&52.68&51.72\\
M\_total       &   6.70 & 11.11 & 15.13  & 8.24 & 5.17 & 9.96 & 5.75 & 12.84 & 15.13 & 9.58&4.98&9.00\\
\midrule
A\_Q1              &   65.13 & 55.17 & 65.71  & 66.86 & 54.98 & 65.33 & 64.75 & 64.94 & 67.05 & 67.62&56.9&68.01\\
A\_Q2              &    76.44 & 73.37 & 65.71  & 75.86 & 73.37 & 72.80 & 74.52 & 71.65 & 68.20 & 71.84&64.94&72.41\\
A\_Q2\_1          &   28.44 & 23.24 & 24.16  & 28.75 & 22.32 & 24.46 & 29.97 & 16.51 & 23.85 & 22.32&15.29&14.68 \\
A\_Q3               &    0.00 & 35.38 & 36.41  & 35.38 & 24.10 & 30.77 & 34.87 & 36.92 & 38.46 & 27.69&6.15&38.97\\
TF Total         &    44.44 & 42.72 & 40.04 & 45.49 & 42.15 & 42.91 & 42.72 & 44.44 & 43.30 & 51.15 &37.36&50.00\\
A\_total       &    17.05 & 16.09 & 13.79  & 18.39 & 15.52 & 12.64 & 15.52 & 12.26 & 16.09 &13.03 &6.90 &12.26\\
\textbf{Overall Acc} &  4.02 &  0.96 &  4.41  & 2.49 & 0.57 & 1.53 & 3.07 & 1.92 & 3.26 &1.72 & 0.77&0.77\\
\midrule
\multicolumn{7}{l}{\textbf{B. Coefficient Similarity to Human (Group-level)}} \\
CSA       
&     0.72   &  0.86   & 0.76  & 0.86  & 0.83  & 0.86 & 0.83 & 0.69 & 0.69 & 0.72 & 0.86 & 0.86 \\
SSA
&    0.93   &  0.90   & 0.86 &  0.86 & 0.97 & 0.90 & 0.90 & 0.86 & 0.86 & 0.86 & 0.90 & 0.86 \\
Jaccard  &    0.67   &  0.57   & 0.20  &  0.20 & 0.80 & 0.57 & 0.50 & 0.20 & 0.33 & 0.20 & 0.50 & 0.20\\
\bottomrule


 \multicolumn{13}{l}{\textbf{Storytelling Format}} \\


\multicolumn{7}{l}{\textbf{A. Accuracy by Question (Individual-level)}} \\
M\_Q1               &   59.58  &   59.96  &  59.77  &  59.58  & 44.06  & 61.69 & 59.96 & 60.54 & 59.96 \\
M\_Q2               &   69.16  &   78.35  &  78.54  &  79.69  & 78.35 & 79.31 & 75.10 & 78.54 & 80.08 \\
M\_Q2.1          &  16.38   &  21.52  & 33.74  & 25.92  & 17.85  &  22.98 & 18.83 & 33.74 & 33.99 \\
M\_Q3    &  31.86  & 9.73 & 36.28 & 31.86 & 9.73 &  30.97 & 31.86 & 27.43 & 27.43\\
TF Total     &  52.87  & 52.49 & 53.07 & 53.26 & 28.35 & 53.45 & 53.64 & 48.85 & 53.26\\
M1\_total       &  3.07  & 7.09 & 15.71 & 9.77 & 9.58 & 8.81 & 5.56 & 16.67 & 16.28 \\
\cmidrule{1-10}
A\_Q1              &  59.96  & 54.79 & 62.45 & 65.52 & 44.64 & 65.52 & 61.30 & 58.81 & 62.07 \\
A\_Q2              &  74.52  & 74.33 & 68.97 & 73.56 & 73.37  & 74.41 & 74.52 & 70.88 & 69.54 \\
A\_Q2.1          &  27.22  & 19.27 & 26.61 & 24.46 & 24.46 & 24.77 & 28.75 & 20.49 & 21.41 \\
A\_Q3               &  36.41  & 24.10 & 36.41 & 36.41 & 26.15 & 27.69 & 33.33 & 34.36 & 38.86 \\
TF Total         &  42.34  & 42.53 & 46.36 & 44.83 & 39.46 & 45.98 & 42.15 & 41.57 & 44.83 \\
A2\_total       &  16.86  & 13.98 & 16.67 & 17.24 & 12.64 & 15.13 & 16.28 & 12.45 & 15.13 \\
\textbf{Overall Acc} &  0.19  & 0.38 & 1.34 & 1.34 & 0.57 & 1.34 & 0.96 & 1.92 & 1.72 \\
\cmidrule{1-10}
\multicolumn{7}{l}{\textbf{B. Coefficient Similarity to Human (Group-level)}} \\
CSA       
&  0.72  & 0.69  & 0.76 & 0.69 & 0.93 & 0.62 & 0.83 & 0.83 & 0.86 \\
SSA
&  0.93  & 0.93 & 0.93 & 0.93 & 0.93 & 0.83 & 0.90 & 0.90 & 0.86 \\
Jaccard  &  0.6  & 0.67 & 0.67 & 0.6 & 0.67 & 0.17 & 0.4 & 0.5 & 0.4 \\
\bottomrule

\end{tabular}
\caption{LLM performance across prompting strategies and interaction formats for music performance (M) and art (A) tasks.}
\label{tab:rq2_full}
\end{table*}

\section{RQ2 : Prompt Injection Methology}
While RQ1 demonstrated LLMs' potential to reproduce human-like choices, it revealed limitations in accurately predicting individual-level. The improved performance in the Human-guided condition suggests that persona presentation and prompt structure significantly impact LLM prediction performance. Therefore, RQ2 systematically analyzes how different persona composition methods and prompt engineering affect LLMs' human behavior prediction.

\subsection{Experiment Setup}
Based on RQ1 results showing superior performance of Human-guided condition, RQ2 adopted this as the baseline setting for testing various persona composition and injection methods. All experimental conditions remained identical to RQ1, including models, API settings, and temperature parameters.
The experiment examined two main axes. First, \textit{Persona Format} included Survey Format (presenting the 65 collected persona items in structured form) and Storytelling Format (reconstructing the same information into biographical narratives). Second, \textit{Prompting Methods} included na\"ive prompts (Base), CoT, RAG, and Few-shot Prompting (providing three response examples from similar personas). This resulted in seven experimental combinations: four techniques for Survey and three for Storytelling.

\subsection{Result and Discussion}
The RQ2 experimental results reveal several key patterns according to different persona format and prompting methods. For detailed results, refer to Table \ref{tab:rq2_full}.

\paragraph{Effect of Persona Format}
Analysis of the overall results reveals that Survey Format consistently outperformed Storytelling Format across most conditions. Specifically, Survey Format demonstrated higher Overall Accuracy than Storytelling Format across the majority of models and prompting methods. For instance, GPT-4o recorded 4.02\% in Survey Format compared to 0.19\% in storytelling Format under Base conditions, while Qwen showed a substantial difference with 4.41\% versus 1.34\%, respectively. These findings suggest that structured information presentation better supports LLMs' ability in information processing.

At the individual item level, Survey Format exhibited marginally higher accuracy on binary questions compared to Storytelling Format. GPT-4o demonstrated slightly superior performance in Survey Format for \texttt{Q2} (70.50\%, 76.44\%) relative to Storytelling Format (69.16\% and 74.52\%, respectively). Despite these modest differences at the item level, the formats produced substantial disparities in Overall Accuracy, indicating that cumulative effects across multiple decision points significantly amplify performance differences.

We suspect that the different Persona Format may hinder clarity of persona items. Specifically, Survey Format appears to facilitate LLMs' ability to understand persona information and establish clear connections to decision-making processes. In contrast, Storytelling Format may introduce attentional dispersion, diverting focus from decision-relevant information. This effect was particularly pronounced in short-answer response items, where GPT-4o showed comparable performance between Survey Format (\texttt{M\_Q2.1}: 15.65\%) and Storytelling Format (16.38\%), yet differed substantially in overall consistency. 


\paragraph{Effect of Prompting Methods}

The most notable finding is that prompting methods demonstrated only limited improvements compared to the Base approach. In the Survey Format, Few-shot prompting yielded mixed results: while it improved accuracy on specific items (GPT-4o showed improvement from 15.65\% to 20.05\% on \texttt{M\_Q2.1} from Base to Few-shot), Overall Accuracy decreased in most cases (Base 4.02\% vs Few-shot 1.72\%). This suggests that providing additional examples may enhance performance on zero-shot Base method but fails to translate into comprehensive prediction accuracy across the complete decision-making process. Similarly, CoT and RAG techniques failed to provide consistent improvements. CoT occasionally resulted in performance degradation (GPT-4o Overall accuracy declined from Base 4.02\% to CoT 2.49\%), indicating that explicit reasoning steps may introduce additional sources of error rather than enhancing prediction capability. RAG maintained intermediate performance levels but showed no clear enhancement over the Base approach.

These findings collectively demonstrate that current prompting methods exhibit limited effectiveness in complex human behavior prediction tasks, suggesting that the challenge of persona-based behavioral simulation may require fundamentally different approaches beyond conventional prompting method.

\paragraph{Note on low Jaccard index}
The most critical finding is the consistently low Jaccard Index across both formats. Survey Format showed Jaccard scores ranging from 0.2-0.67, while Storytelling Format exhibited similar levels at 0.17-0.67. The Storytelling Format was originally designed to integrate persona information as a narrative, encouraging LLMs to consider comprehensive character backgrounds rather than isolated attributes. The intention was to prevent excessive focus on specific variables and promote more balanced persona understanding. However, the consistently low Jaccard Index across both formats suggests that LLMs possess fundamentally different abilities in variable selection and weighting compared to humans.

This finding extends beyond information presentation methods, revealing intrinsic limitations in current LLM architectures; they have low ability to understand and reproduce the complex and contextual characteristics of human decision-making processes. Regardless of whether information is presented in structured or narrative form, LLMs heavily rely on limited and biased portion of the given persona, exposing the inherent constraints of current technology in persona-based behavioral simulation. While LLMs can capture directional tendencies and statistical significance at the group level to some extent, their ability to identify decision-making factors that humans actually consider important remains unimproved despite modifications in persona format and prompting methods.
So, the challenge of achieving human-like behavioral simulation may require fundamental advances in model architecture and training paradigms, rather than incremental improvements in persona format or prompting method. The heavy reliance on limited portion of persona indicates that current LLMs may lack the human-like mechanisms necessary for authentic persona-based decision modeling, highlighting a crucial area for future research in developing more sophisticated approaches to human behavior simulation.
\section{Conclusion}
This study presents the first comprehensive evaluation of LLMs' ability to simulate individual human economic decision-making using real 522 human persona data in Pay-What-You-Want (PWYW) pricing experiments. Our findings reveal that while LLMs demonstrate reasonable group-level tendencies, they struggle significantly with precise individual-level accuracy, achieving overall accuracy rates below 5\% across all tested models and conditions. Notably, commonly adopted advanced prompting method such as CoT, RAG, and Few-shot prompting showed no substantial improvements over na\"ive baseline methods, and structured survey format consistently outperformed narrative storytelling approaches. The consistently low Jaccard indices across all experimental conditions suggest fundamental limitations in current LLM architectures' ability to identify and weigh decision-making factors that humans actually consider important in their decision making.

These results provide crucial empirical evidence for the current boundaries of LLM-based human behavior simulation and highlight the need for fundamental advances in model architecture and training paradigms rather than incremental improvements in prompt engineering. However, this study has two limitations. First, we did not conduct detailed analysis of which specific persona components among the 65 collected items most significantly influence prediction performance. Second, we did not explore advanced technical modeling approaches such as ensemble methods or fine-tuning that could potentially improve accuracy. Despite these two limitations, this work establishes a rigorous methodological framework for evaluating LLM's ability in persona-based simulation and emphasizes the importance of realistic expectations for current LLM technology in individual human behavior modeling.

\bibliography{custom}

\newpage

\newpage

\setcounter{secnumdepth}{2}

\appendix

\setcounter{secnumdepth}{2}
\section{Detailed Persona Information}
This section provides the full list of 65 items in persona questionnaire collected from participants, categorized into seven thematic groups. Each item is presented in English translation first, followed by its original Korean version.

\subsection{Basic Demographic Information (11 items)}
\label{app:persona.demographic}

\begin{enumerate}[label=\textbf{Q\arabic*.}]

    \item What is your gender?\\
    \textbf{성별은 무엇입니까?}\\
    ① Male (남자) \hspace{1em} ② Female (여자)\\

    \item What is your age?\\
    \textbf{연령은 어떻게 되십니까?}\\
    Age: \underline{\hspace{0.7cm}} (만 \underline{\hspace{0.7cm}} 세)
    \\
    \item What is your marital status?\\
    \textbf{결혼 유무는 어떻게 되십니까?}\\
    ① Married (기혼) \\
    ② Single (미혼) \\
    ③ Divorced, Bereaved, etc. (이혼, 사별 등)\\

    \item How many people currently live with you (including yourself)?\\
    \textbf{현재 함께 거주하고 있는 가구원 수는 몇 명입니까? (본인을 포함)}\\
    Number of household members: \underline{\hspace{0.7cm}} (\underline{\hspace{0.7cm}} 명)\\

    \item Who do you live with? (Select all that apply)\\
    \textbf{귀하의 가구 구성은 어떻게 되십니까? (*중복 응답)}
    \begin{itemize}[leftmargin=1em]
        \item[①] Grandparent(s) (조부 또는 조모) 
        \item[②] Parent(s) (부 또는 모) 
        \item[③] Sibling(s) (형제/자매) 
        \item[④] Relative(s) (친척) 
        \item[⑤] Spouse (배우자) 
        \item[⑥] Son(s)/Daughter(s) (아들/딸) 
        \item[⑦] Alone (혼자) 
        \item[⑧] Others (기타)\\
    \end{itemize}

    \item What is your education level?\\
    \textbf{학력은 어떻게 되십니까?}\\
    ① Middle school or less (중학교 졸업 이하) → Q7\\
    ② High school graduate (고등학교 졸업) → Q7\\
    ③ Currently in university (대학교 재학) → Q6-1\\
    ④ University graduate (대학교 졸업) → Q6-1\\
    ⑤ Currently in graduate school (대학원 재학) → Q6-1\\
    ⑥ Graduate school graduate (대학원 졸업) → Q6-1

    \item[\textbf{Q6-1.}] What is your university major?\\
    \textbf{대학교 전공은 어떻게 되십니까?}\\
    ① Humanities (인문계열)  \\
    ② Social sciences (사회계열) \\
    ③ Business (경상계열) \\
    ④ Education (교육계열)\\
    ⑤ Engineering (공학계열) \\
    ⑥ Natural sciences (자연계열)\\
    ⑦ Arts/Physical Education (예체능계열) \\
    ⑧ Other (기타)\\

    \item What is your occupation?\\
    \textbf{직업은 어떻게 되십니까?}\\
    ① Office worker (사무직) \\
    ② Civil servant (공무원) \\
    ③ Professional (전문직)\\
    ④ Sales/service (판매/서비스직) \\
    ⑤ Executive/Assembly member (고위임원/의회의원)\\
    ⑥ Agriculture/Fishery worker (농업/어업 종사자) \\
    ⑦ Technician (기술공)\\
    ⑧ Manual labor (단순노무) \\
    ⑨ Self-employed (자영업)\\
    ⑩ Housewife (전업주부) \\
    ⑪ Other (기타)\\

    \item What is your father's education level?\\
    \textbf{아버지의 학력은 어떻게 되십니까?}\\
    ① Middle school or less (중학교 졸업 이하)\\
    ② High school graduate (고등학교 졸업)\\
    ③ University graduate (대학교 졸업)\\
    ④ Graduate school graduate (대학원 졸업)\\

    \item What is your mother's education level?\\
    \textbf{어머니의 학력은 어떻게 되십니까?}\\
    ① Middle school or less (중학교 졸업 이하)\\
    ② High school graduate (고등학교 졸업)\\
    ③ University graduate (대학교 졸업)\\
    ④ Graduate school graduate (대학원 졸업)\\

    \item What is your average monthly personal income (before taxes and insurance)?\\
    \textbf{지난 1년간 세금 및 보험을 제하기 전 월 평균 개인소득은 어떻게 되십니까? (가구 전체 합산 아님)}\\
    ① Not applicable (해당 없음) \\
    ② Less than 1 million won (100만원 미만)\\
    ③ 1–2 million won (100–200만원) \\
    ④ 2–3 million won (200–300만원)\\
    ⑤ 3–4 million won (300–400만원) \\
    ⑥ 4–5 million won (400–500만원)\\
    ⑦ 5–6 million won (500–600만원) \\
    ⑧ 6–7 million won (600–700만원)\\
    ⑨ Over 7 million won (700만원 이상)\\

    \item What is your average monthly personal expenditure?\\
    \textbf{월 평균 개인 지출은 어떻게 되십니까?}\\
    ※ Excludes fixed expenses such as rent, taxes, vehicle maintenance, and transportation. Includes food, shopping, cultural and leisure expenses.\\
    ※ 월세, 세금, 차량유지비, 교통비 등 고정비 제외. 식비, 쇼핑, 문화 및 여가비 포함.\\
    ① Less than 500,000 won (50만원 미만)\\
    ② 500,000–1 million won (50–100만원 미만)\\
    ③ 1–1.5 million won (100–150만원 미만)\\
    ④ 1.5–2 million won (150–200만원 미만)\\
    ⑤ 2–2.5 million won (200–250만원 미만)\\
    ⑥ 2.5–3 million won (250–300만원 미만)\\
    ⑦ 3–3.5 million won (300–350만원 미만)\\
    ⑧ 3.5–4 million won (350–400만원 미만)\\
    ⑨ 4–4.5 million won (400–450만원 미만)\\
    ⑩ 4.5–5 million won (450–500만원 미만)\\
    ⑪ Over 5 million won (500만원 이상)

\end{enumerate}

\subsection{Cultural Participation (6 items)}
\label{app:persona.cult-part}

In the past year (from September 1, 2020 to August 31, 2021), have you participated in the following activities? If yes, how many times?\\
\textbf{귀하는 지난 1년간 (2020년 9월 1일–2021년 8월 31일) 다음의 활동을 하신 적이 있습니까? 하셨다면 얼마나 하셨습니까?}

\vspace{1em}

\begin{enumerate}[label=\textbf{Q\arabic*.}]
    \item Attended an art exhibition\\
    \textbf{미술 전시회 관람} \\
    ① No (없다) \hspace{1em} ② Yes (있다): ( \underline{\hspace{0.7cm}} ) times (회)\\

    \item Attended a classical music concert or opera\\
    \textbf{클래식 음악회, 오페라 공연 관람} \\
    ① No (없다) \hspace{1em} ② Yes (있다): ( \underline{\hspace{0.7cm}} ) times (회)\\

    \item Read literary works\\
    \textbf{문학작품 읽기} \\
    ① No (없다) \hspace{1em} ② Yes (있다): ( \underline{\hspace{0.7cm}} ) times (회)\\

    \item Attended a play\\
    \textbf{연극 관람} \\
    ① No (없다) \hspace{1em} ② Yes (있다): ( \underline{\hspace{0.7cm}} ) times (회)\\

    \item Attended a musical\\
    \textbf{뮤지컬 관람} \\
    ① No (없다) \hspace{1em} ② Yes (있다): ( \underline{\hspace{0.7cm}} ) times (회)\\

    \item Attended a dance performance (ballet, modern dance, Korean traditional dance)\\
    \textbf{무용 공연 관람 – 발레, 현대무용, 한국무용} \\
    ① No (없다) \hspace{1em} ② Yes (있다): ( \underline{\hspace{0.7cm}} ) times (회)\\
\end{enumerate}

\subsection{Cultural Arts Attitudes (19 items)}
\label{app:persona.cult-att}

Please respond based on your usual thoughts.\\
\textbf{평소 귀하께서 생각하시는 것을 토대로 응답해주십시오.}

\vspace{1em}

\noindent Response scale: ① Strongly Disagree (전혀 그렇지 않다), ② Disagree (그렇지 않다), ③ Neutral (보통이다), ④ Agree (그렇다), ⑤ Strongly Agree (매우 그렇다)

\vspace{1em}

\begin{enumerate}[label=\textbf{Q\arabic*.}]
    \setcounter{enumi}{6}  

    \item I enjoy beautiful things.\\
    \textbf{나는 아름다운 것들을 즐긴다.}\\

    \item I believe having good manners is important.\\
    \textbf{나는 좋은 매너를 갖는 것이 필요하다고 생각한다.}\\

    \item I believe culture is more important than material wealth.\\
    \textbf{나는 물질적 부보다 문화가 더 중요하다고 생각한다.}\\

    \item I enjoy things related to cultural arts.\\
    \textbf{나는 문화예술과 관련된 것들을 즐긴다.}\\

    \item I consider myself cultured in terms of cultural arts.\\
    \textbf{나는 문화예술 측면에서 교양있는 사람이라고 생각한다.}\\

    \item People think I have a taste for consuming cultural arts.\\
    \textbf{사람들은 나를 문화예술을 소비하는 취향을 가진 사람으로 생각한다.}\\

    \item I try to participate in activities related to cultural arts.\\
    \textbf{나는 문화예술과 관련된 활동에 참여하려고 노력한다.}\\

    \item I try to develop tastes related to cultural arts.\\
    \textbf{나는 문화예술과 관련된 취향을 가지려고 노력한다.}\\

    \item I try to refine myself by consuming cultural arts.\\
    \textbf{나는 문화예술을 소비함으로써 고상해지려고 노력한다.}\\

    \item I prefer classical music performances.\\
    \textbf{나는 클래식 공연 관람을 선호한다.}\\

    \item I prefer opera performances.\\
    \textbf{나는 오페라 공연 관람을 선호한다.}\\

    \item I prefer dance performances.\\
    \textbf{나는 무용 공연 관람을 선호한다.}\\

    \item I prefer art exhibitions.\\
    \textbf{나는 미술 전시회 관람을 선호한다.}\\

    \item I prefer traditional music performances.\\
    \textbf{나는 전통 음악 공연 관람을 선호한다.}\\

    \item I have a positive view of classical musicians’ professions.\\
    \textbf{나는 클래식 음악 연주자의 직업을 긍정적으로 생각한다.}\\

    \item I have a positive view of vocal artists such as opera singers.\\
    \textbf{나는 오페라와 같은 성악가의 직업을 긍정적으로 생각한다.}\\

    \item I have a positive view of dancers’ professions.\\
    \textbf{나는 무용가의 직업을 긍정적으로 생각한다.}\\

    \item I have a positive view of artists’ professions.\\
    \textbf{나는 미술가의 직업을 긍정적으로 생각한다.}\\

    \item I have a positive view of traditional musicians’ professions.\\
    \textbf{나는 전통 음악가의 직업을 긍정적으로 생각한다.}\\

\end{enumerate}

\subsection{Cultural Arts Knowledge Level (4 items)}
\label{app:persona.cult-know}

Please answer based on your current knowledge. These questions are meant to assess your level of cultural arts knowledge.\\
\textbf{문화예술 지식을 측정하기 위한 질문이니 정확한 측정을 위해 평소 아시는 것을 토대로 응답해주십시오.}

\vspace{1em}

\begin{enumerate}[label=\textbf{Q\arabic*.}]
    \setcounter{enumi}{25}  

    \item Who is the composer of the classical piece, Symphony “Fate”?\\
    \textbf{클래식 곡, 교향곡 ‘운명’의 작곡가는 누구입니까?}\\
    ① Mozart (모차르트) \\
    ② Bach (바흐) \\
    ③ Beethoven (베토벤)\\
    ④ Chopin (쇼팽)\\
    ⑤ I don't know (모름)\\

    \item Who is the composer of the opera “Rigoletto”?\\
    \textbf{클래식 곡, 오페라 ‘리골레토’의 작곡가는 누구입니까?}\\
    ① Mozart (모차르트) \\ ② Verdi (베르디) \\ ③ Puccini (푸치니)\\
    ④ Wagner (바그너) \\ ⑤ I don't know (모름)\\

    \item Who is the painter of the artwork “Mona Lisa”?\\
    \textbf{미술작품 ‘모나리자’의 작가는 누구입니까?}\\
    ① Van Gogh (반 고흐) \\ ② Claude Monet (클로드 모네) \\ ③ Pablo Picasso (파블로 피카소)\\
    ④ Leonardo da Vinci (레오나르도 다빈치) \\ ⑤ I don't know (모름)\\

    \item Who is the painter of the artwork “The Scream”?\\
    \textbf{미술작품 ‘절규’의 작가는 누구입니까?}\\
    ① Edvard Munch (에드바르 뭉크) \\ ② Paul Gauguin (폴 고갱) \\ ③ Edgar Degas (에드가 드가)\\
    ④ Paul Cézanne (폴 세잔) \\ ⑤ I don't know (모름)\\

\end{enumerate}

\subsection{Cultural Education Experience (9 items)}
\label{app:persona.cult-ed}

Please answer the following questions regarding your experience with cultural and arts education.\\
\textbf{다음의 설문에 응답해 주십시오.}

\vspace{1em}

\begin{enumerate}[label=\textbf{Q\arabic*.}]
    \item Before age 18, did you receive education related to classical music, opera, or musical theatre at school (elementary, middle, high school)?\\
    \textbf{만 18세 이전 학교내 (초, 중, 고등학교)에서 클래식음악, 오페라, 뮤지컬과 관련된 교육을 받은 경험이 있으십니까?}\\
    ① No (없다) \hspace{2em} ② Yes (있다)\\

    \item Before age 18, did you receive education related to art or exhibitions at school?\\
    \textbf{만 18세 이전 학교내에서 미술, 전시와 관련된 교육을 받은 경험이 있으십니까?}\\
    ① No (없다) \hspace{2em} ② Yes (있다)\\

    \item Before age 18, did you receive education related to classical music, opera, or musicals outside of school (e.g., private institutes, cultural centers, private lessons, clubs)?\\
    \textbf{만 18세 이전 학교외에서 클래식음악, 오페라, 뮤지컬과 관련된 교육을 받은 경험이 있으십니까? (학교외는 사설 학원, 문화센터, 개인레슨, 동호인 모임 등을 의미합니다.)}\\
    ① No (없다) \hspace{2em} ② Yes (있다)\\

    \item Before age 18, did you receive art or exhibition-related education outside of school?\\
    \textbf{만 18세 이전 학교외에서 미술, 전시와 관련된 교육을 받은 경험이 있으십니까? (학교외는 사설 학원, 문화센터, 개인레슨, 동호인 모임 등을 의미합니다.)}\\
    ① No (없다) \hspace{2em} ② Yes (있다)\\

    \item After age 19 and before September 1, 2020, did you receive classical music, opera, or musical education outside of school?\\
    \textbf{만 19세 이후부터 과거 1년 전까지 (2020년 9월 1일) 학교외에서 클래식음악, 오페라, 뮤지컬과 관련된 교육을 받은 경험이 있으십니까? (학교외는 사설 학원, 문화센터, 개인레슨, 동호인 모임 등을 의미합니다.)}\\
    ① No (없다) \hspace{2em} ② Yes (있다)\\

    \item After age 19 and before September 1, 2020, did you receive art or exhibition-related education outside of school?\\
    \textbf{만 19세 이후부터 과거 1년 전까지 (2020년 9월 1일) 학교외에서 미술, 전시와 관련된 교육을 받은 경험이 있으십니까? (학교외는 사설 학원, 온라인 교육, 과외, 방송 교육 등을 의미합니다.)}\\
    ① No (없다) \hspace{2em} ② Yes (있다)\\

    \item Between September 1, 2020 and August 31, 2021, did you receive classical music, opera, or musical education outside of school?\\
    \textbf{지난 1년 전부터 현재까지 (2020년 9월 1일–2021년 8월 31일) 학교외에서 클래식음악, 오페라, 뮤지컬 연관된 교육을 받은 경험이 있으십니까? (학교외는 사설 학원, 온라인 교육, 과외, 방송 교육 등을 의미합니다.)}\\
    ① No (아니오) \hspace{2em} ② Yes (예)\\

    \item Between September 1, 2020 and August 31, 2021, did you receive art or exhibition-related education outside of school?\\
    \textbf{지난 1년 전부터 현재까지 (2020년 9월 1일–2021년 8월 31일) 학교외에서 미술, 전시와 관련된 교육을 받은 경험이 있으십니까? (학교외는 사설 학원, 온라인 교육, 과외, 방송 교육 등을 의미합니다.)}\\
    ① No (아니오) \hspace{2em} ② Yes (예)\\

    \item Did you graduate from an arts high school majoring in art or music, or did you major in an art- or music-related field at university?\\
    \textbf{예술계 고등학교에서 미술 또는 음악 분야를 전공으로 졸업하거나, 대학교에서 미술 또는 음악 관련 학과를 전공하셨습니까?}\\
    Art-related majors: Painting, Photography, Sculpture, Craft, Cartoon, Design, etc. \\
    - \textbf{미술 분야 전공:} 회화, 사진, 조각, 공예, 만화, 디자인 등 \\
    Music-related majors: Classical music, Korean traditional music, popular music, etc\\
    - \textbf{음악 분야 전공:} 클래식음악, 국악, 대중음악 등 \\
    ① No (아니오) \hspace{2em} ② Yes (예)\\
\end{enumerate}

\subsection{Institution Satisfaction (4 items)}
\label{app:persona.inst}

\begin{enumerate}[label=\textbf{Q\arabic*.}]
    \setcounter{enumi}{29}
    
    \item Have you ever visited the National Museum of Korea?\\
    \textbf{귀하께서는 \textbf{국립중앙박물관}에 방문하신 경험이 있습니까?} \\
    ① No (아니오) → Skip to Q31\\
    ② Yes (예) → Go to Q30-1\\

    \item[\textbf{Q30-1.}] How satisfied were you overall with your visit to the National Museum of Korea?\\
    \textbf{귀하께서 국립중앙박물관에 방문 후 느꼈던 \textbf{전반적인 만족도}는 어떠십니까?} \\
    ① Very dissatisfied (매우 불만족스럽다) \\
    ② Dissatisfied (불만족스럽다) \\
    ③ Neutral (보통이다) \\
    ④ Satisfied (만족스럽다) \\
    ⑤ Very satisfied (매우 만족스럽다)\\

    \item Have you ever visited the Seoul Arts Center (예술의전당)?\\
    \textbf{귀하께서는 \textbf{예술의전당}에 방문하신 경험이 있습니까?} \\
    ① No (아니오) → Skip to Section VI \\
    ② Yes (예) → Go to Q31-1\\

    \item[\textbf{Q31-1.}] How satisfied were you overall with your visit to the Seoul Arts Center?\\
    \textbf{귀하께서 예술의전당에 방문 후 느꼈던 \textbf{전반적인 만족도}는 어떠십니까?} \\
    ① Very dissatisfied (매우 불만족스럽다) \\
    ② Dissatisfied (불만족스럽다) \\
    ③ Neutral (보통이다) \\
    ④ Satisfied (만족스럽다) \\
    ⑤ Very satisfied (매우 만족스럽다) \\
\end{enumerate}

\subsection{Social Norms and Attitudes (10 items)}
\label{app:persona.social}

Please respond based on your usual thoughts.\\
\textbf{평소 귀하께서 생각하시는 것을 토대로 응답해주십시오.}

\vspace{0.5em}

\noindent Response scale: ① Strongly Disagree (전혀 그렇지 않다), ② Disagree (그렇지 않다), ③ Neutral (보통이다), ④ Agree (그렇다), ⑤ Strongly Agree (매우 그렇다)

\vspace{1em}

\begin{enumerate}[label=\textbf{Q\arabic*.}]
    \item I feel good when I act according to my conscience.\\
    \textbf{나는 양심에 따라 행동을 할 때 기분이 좋다.}\\

    \item I feel guilty when I pay less than an appropriate amount.\\
    \textbf{나는 적정금액보다 낮은 금액을 지불했을 때 양심에 걸린다.}\\

    \item I feel a moral obligation to buy cultural products (e.g., performances, exhibitions) even if they are expensive, if it is for charity.\\
    \textbf{나는 자선단체 기부를 위해서라면 문화예술상품(공연, 전시 등)이 비싸더라도 구매해야 한다는 도덕적 의무감을 느낀다.}\\

    \item I feel a moral obligation to attend cultural events (e.g., performances, exhibitions) for charity even if I am busy.\\
    \textbf{나는 자선단체 기부를 위해서라면 문화예술상품(공연, 전시 등)에 가려고 노력해야 한다는 도덕적 의무감을 느낀다.}\\

    \item If the quality of cultural products (e.g., performances, exhibitions) is good, I don't mind paying more.\\
    \textbf{나는 문화예술상품(공연, 전시 등)의 질이 좋다면 더 많은 돈을 지불할지라도 개의치 않는다.}\\

    \item People around me (family, friends, neighbors) will understand if I pay more to attend cultural events for charity.\\
    \textbf{나의 주변 사람들(가족, 친구, 이웃)은 자선단체 기부를 위해서라면 문화예술상품(공연, 전시 등) 관람에 더 많은 돈을 지불하는 나의 행동에 공감할 것이다.}\\

    \item Other people's attitudes and actions influence me.\\
    \textbf{다른 사람들의 태도나 행동이 나에게 영향을 미친다.}\\

    \item I sometimes act against my will because I am conscious of others’ criticism.\\
    \textbf{다른 사람의 비난을 의식하여 원치 않는 행동을 하는 경우가 있다.}\\

    \item I am willing to pay the same amount that others paid, even if I would normally pay less.\\
    \textbf{나는 사람들의 눈을 의식하여 다른 사람이 지불한 금액을 기꺼이 따라 지불할 용의가 있다.}\\

    \item I am willing to pay more if others around me also pay a higher amount.\\
    \textbf{다른 사람들의 눈을 의식하여 다른 사람이 높은 비용을 지불하더라도 기꺼이 따라할 용의가 있다.}\\
\end{enumerate}

\section{Detailed Question Form}
\begin{figure*}[t]
\centering
\begin{minipage}{0.95\textwidth}
\small

\begin{center}
\textbf{CASE 1}
\end{center}

\vspace{1em}

\textbf{I. 다음은 국립중앙박물관에 대한 설명입니다.} \\
\textbf{I. The following is an introduction to the National Museum of Korea.}

\vspace{1em}

국립중앙박물관은 \textbf{우리나라를 대표하는 국립박물관}으로  
1915년 12월 1일 경복궁에서 개관한 조선총독부박물관을 인수하여 1945년 12월 3일 \textit{국립박물관}이라는 명칭으로 개관하였습니다.  
이후 1972년 7월 \textit{국립중앙박물관}으로 명칭이 변경되었고, 현재 위치인 용산으로 2005년 이전하였습니다.  
국립중앙박물관은 과거로부터 이어온 세계 유산을 수집·보존·전시하고 있으며,  
특히 우리 문화의 보급을 통해 세계와 소통하려 노력하고 있습니다. \\

The National Museum of Korea, the \textbf{representative national museum of South Korea},  
was originally opened as the Government-General Museum of Korea in Gyeongbokgung Palace on December 1, 1915.  
It was reestablished under the name “National Museum” on December 3, 1945.  
The name was changed to “National Museum of Korea” in July 1972, and the museum was relocated to its current site in Yongsan in 2005.  
The museum is dedicated to collecting, preserving, and exhibiting global heritage from the past,  
and especially strives to communicate Korean culture to the world.

\vspace{1em}

국립중앙박물관은 \textbf{무료}로 운영되는 상설전시관, 어린이박물관 등의 시설과  
\textbf{유료}로 운영되는 특별기획전시관을 보유하고 있습니다.  
또한, \textbf{행사, 공연, 교육 등 문화예술 체험 프로그램}을 통해  
성인뿐 아니라 어린이, 외국인 등 다양한 관람객이 참여할 수 있도록 운영하고 있습니다. \\

The museum operates permanent exhibition halls and a children's museum that are \textbf{free of charge},  
along with \textbf{special exhibitions that require paid admission}.  
It also offers \textbf{cultural and artistic programs}, such as events, performances, and educational experiences,  
designed to encourage participation from a wide audience, including adults, children, and foreign visitors.

\vspace{1.5em}

\begin{minipage}[t]{0.48\textwidth}
\centering
\includegraphics[width=\linewidth]{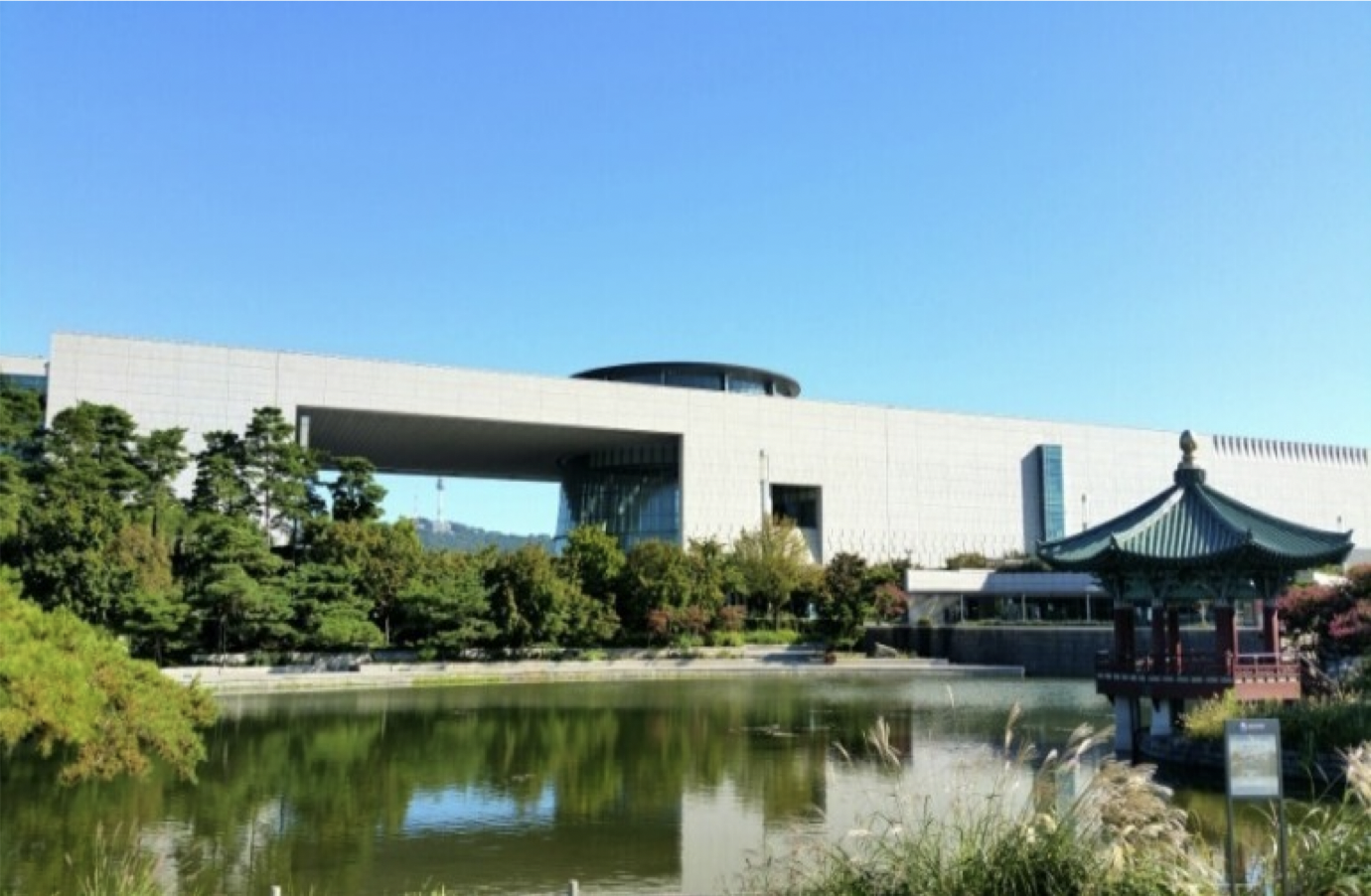} \\[0.5em]
\small [그림 1] 국립중앙박물관 전경 [Figure 1] View of the National Museum of Korea
\end{minipage}
\hfill
\begin{minipage}[t]{0.48\textwidth}
\centering
\includegraphics[width=\linewidth]{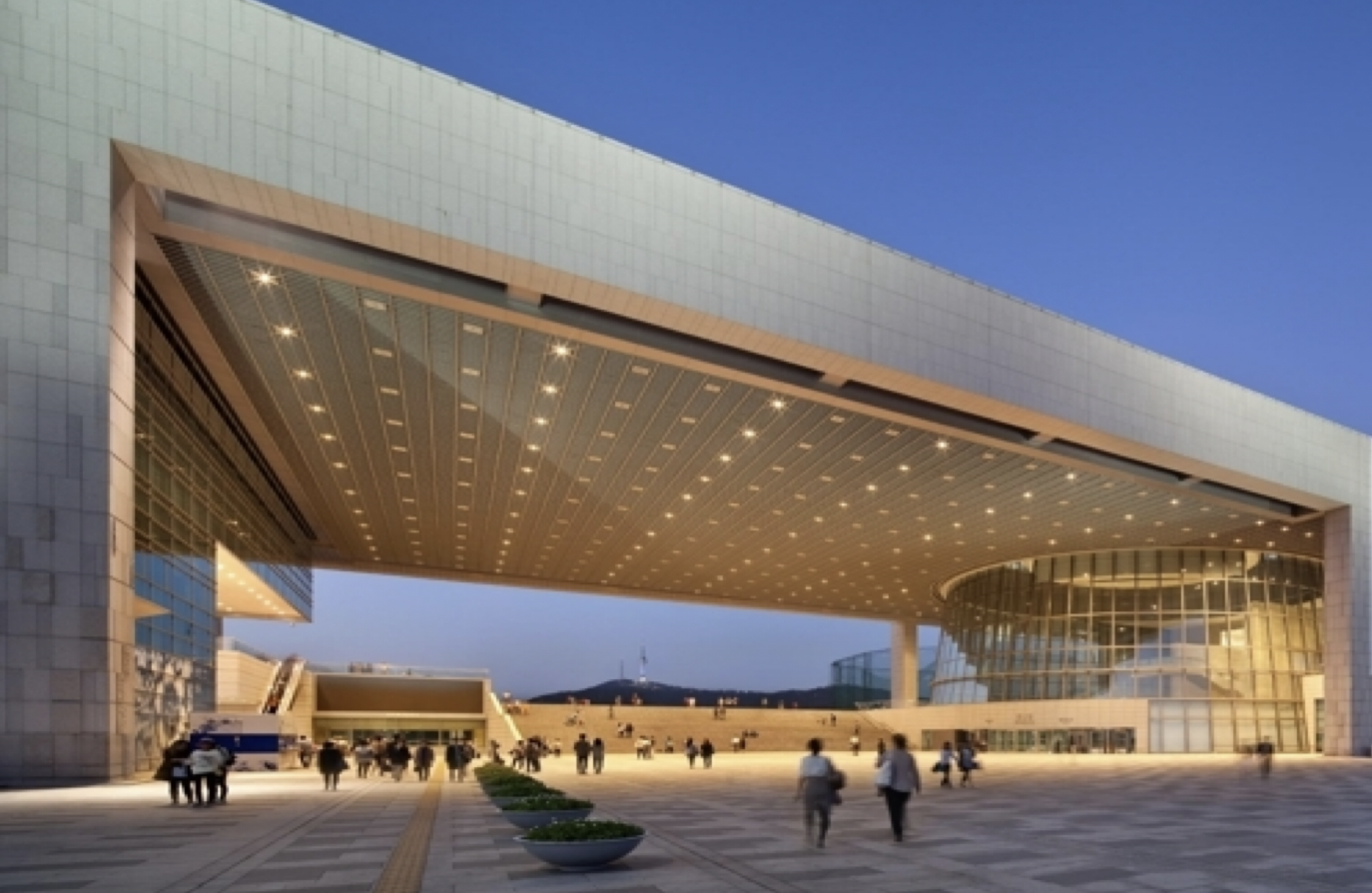} \\[0.5em]
\small [그림 2] 국립중앙박물관 외관 [Figure 2] Exterior of the National Museum of Korea
\end{minipage}

\end{minipage}
\caption{Case 1 (Art) Introduction part of the PWYW questionnaire}
\label{fig:case1_intro}
\end{figure*}

\begin{figure*}[t]
\centering
\begin{minipage}{0.95\textwidth}
\small

\textbf{II. 다음은 전시 방문 가상상황에 대한 설명입니다.} \\
\textbf{II. The following is a hypothetical scenario about visiting an exhibition.}

\vspace{1em}

여러분은 현재 서울 용산구 국립중앙박물관 근처에 있고 3–4시간 정도 여유시간이 있습니다.  
바로 옆 국립중앙박물관에서는 성인기준 \textbf{9,000원}으로  
\textit{《시대의 얼굴, 셰익스피어에서 에드 시런까지 2021》}가 특별전시하고 있습니다. \\

You are currently near the National Museum of Korea in Yongsan, Seoul, and have about 3–4 hours of free time.  
Next door, the National Museum is holding a special exhibition titled  
\textit{“The Face of the Era: From Shakespeare to Ed Sheeran 2021”} with an admission fee of \textbf{₩9,000} for adults.

\vspace{1em}

\begin{center}
\includegraphics[width=0.2\textwidth]{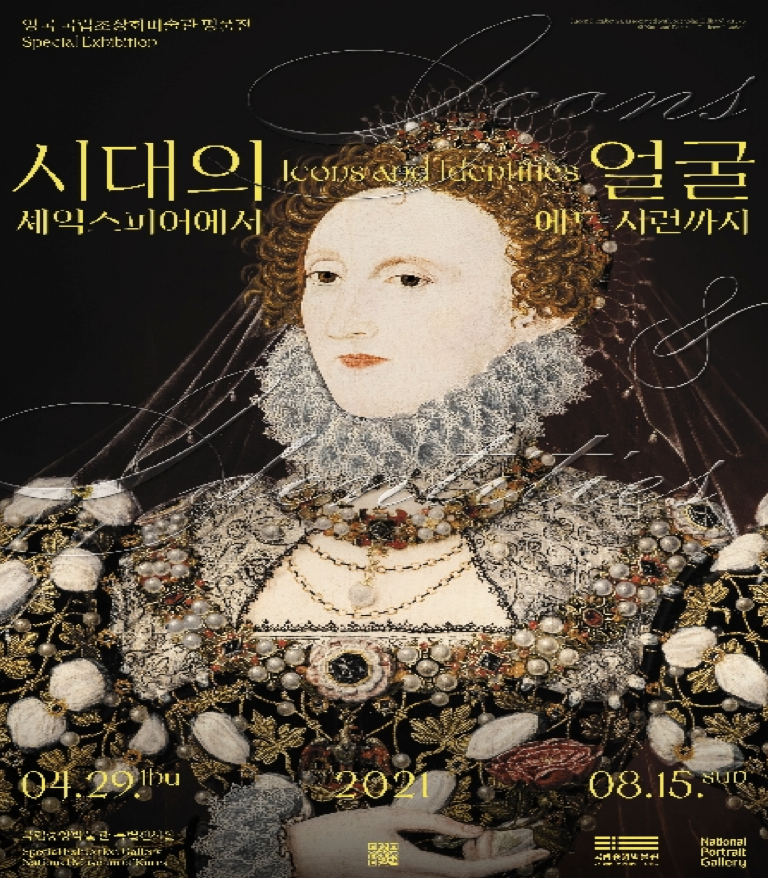} \\
\small [그림] 전시 포스터 / [Figure] Exhibition Poster
\end{center}

\vspace{1em}

\textbf{《전시소개》 Exhibition Description} \\
이번 전시는 ‘76명의 역사적 인물’이 그려진 \textbf{‘초상화’}라는 예술 장르를 중심으로 한 기획전입니다.  
영국 국립초상화박물관의 소장품이 한국에서 처음 공개되며, 세계 역사와 문화를 빛낸 인물들을 만날 수 있습니다.  
세익스피어, 엘리자베스 1세, 뉴턴, 비틀즈, 에드 시런 등이 등장합니다.\\

This exhibition focuses on the genre of \textbf{portraiture}, showcasing depictions of 76 historical figures.  
For the first time, the National Portrait Gallery (UK) brings its collection to Korea.  
Visitors will encounter world-renowned figures such as Shakespeare, Queen Elizabeth I, Newton, The Beatles, and Ed Sheeran.

\vspace{1.5em}

\begin{enumerate}
\item 귀하께서는 \textbf{국립중앙박물관의 특별전시 ‘시대의 얼굴’을 9,000원을 지불하고 관람}할 의도가 있으십니까?\\
\textbf{Would you be willing to pay ₩9,000 to see the special exhibition "The Face of the Era" at the National Museum of Korea?}  
① 아니오 (No) \hfill ② 그렇다 (Yes)

\vspace{1em}

\item 만약 \textbf{위 전시에 대해 방문객이 원하는 가격을 스스로 정해서 지불}할 수 있다면, 귀하께서는 이 경우 방문할 의향이 있으십니까?\\
\textbf{If you could choose your own admission fee for this exhibition, would you still want to visit it?}  
① 아니다 (No) $\rightarrow$ Q3로 이동 (Go to Q3) \\
② 그렇다 (Yes) $\rightarrow$ Q2-1로 이동 (Go to Q2-1)

\vspace{1em}

\item[Q2-1.] \textbf{방문하실 경우 얼마 정도 관람료로 낼 의향이} 있으십니까?\\
\textbf{If you choose to visit, how much would you be willing to pay as admission?}  
\_\_\_\_ 원 (KRW)

\vspace{1em}

\item[Q3.] 귀하께서 \textbf{위 전시에 대해 관람 의사가 없다고 답변하신 이유는 무엇입니까?} (복수선택 가능)\\
\textbf{If you are not willing to attend the exhibition, what are the reasons? (Check all that apply)}  
\begin{itemize}
\item ① 해당 기관에서 제공하는 전시 및 프로그램에 관심이 없다 \\
\hspace{2em} I am not interested in the exhibitions or programs offered at the institution.
\item ② 위 전시에 대한 프로그램이 만족스럽지 않다 \\
\hspace{2em} The content of the exhibition is not satisfactory.
\item ③ 위 전시의 관람 여부를 판단하기 위해 필요한 정보가 충분히 주어지지 않았다 \\
\hspace{2em} There is insufficient information to judge the value of the exhibition.
\item ④ 나는 위의 전시와 관련된 주제에 관심이 없다 \\
\hspace{2em} I am not interested in the theme of the exhibition.
\item ⑤ 나는 상대적으로 다른 여가활동을 더 선호한다 \\
\hspace{2em} I prefer other leisure activities over this exhibition.
\item ⑥ 기타 (직접 서술): \_\_\_\_\_\_\_\_\_\_\_\_\_\_\_ \\
\hspace{2em} Other: \_\_\_\_\_\_\_\_\_\_\_\_\_\_\_
\end{itemize}
\end{enumerate}

\end{minipage}
\caption{Case 1 (Art) Question part of the PWYW questionnaire}
\label{fig:case1_question}
\end{figure*}

\begin{figure*}[t]
\centering
\begin{minipage}{0.95\textwidth}
\small

\begin{center}
\textbf{CASE 2}
\end{center}

\vspace{1em}

\textbf{III. 다음은 예술의전당에 대한 설명입니다.} \\
\textbf{III. The following is an introduction to the Seoul Arts Center.}

\vspace{1em}

예술의전당은 \textbf{우리나라를 대표하는 공공기관 공연장}으로 문화예술의 창달과 진흥,  
문화예술향유기회 확대를 목표로 클래식 음악, 오페라 등의 공연을 비롯하여 미술작품의 수집·전시를 목적으로 1988년 설립되었습니다. \\

The Seoul Arts Center is a \textbf{leading public cultural venue in Korea} established in 1988 with the goal of promoting and advancing the arts.  
It aims to expand public access to cultural experiences, including classical music, opera performances,  
and exhibitions of visual artworks.

\vspace{0.8em}

예술의전당은 2016년 누적 관람객이 5천만 명을 넘었고 음악당, 오페라하우스, 서울서예박물관, 한가람미술관, 한가람디자인미술관 등의 시설을 갖추고 있습니다.  
예술의전당은 관람객에게 \textbf{공연, 전시, 교육 등의 문화예술경험}을 제공하고 있으며 \textbf{서울시 서초구}에 위치한 국가대표 문화예술기관입니다. \\

As of 2016, the center had welcomed over 50 million visitors. It houses multiple facilities such as a concert hall,  
opera house, calligraphy museum, Hangaram Art Museum, and the Design Museum.  
Located in \textbf{Seocho-gu, Seoul}, it offers diverse \textbf{cultural and artistic experiences}, including performances, exhibitions, and educational programs.

\vspace{0.8em}

예술의전당 음악당에서는  
① 예술가나 기획자가 직접 대관하여 진행하는 \textbf{대관공연}과  
② 예술의전당에서 \textbf{특별 기획한 프로그램들}을 소개하고 있으며,  
③ 한화생명, KT, 신세계 등과 같은 기업의 \textbf{후원으로 매달 한 번 오전 11시에 클래식 음악 콘서트}를 제공하고 있습니다. \\

At the Seoul Arts Center’s Music Hall,  
① \textbf{private performances} organized by artists or planners and  
② \textbf{special programs curated by the center} are presented.  
③ With sponsorship from companies such as Hanwha Life, KT, and Shinsegae, the center also offers  
a \textbf{monthly classical music concert at 11 a.m.}.

\vspace{1.5em}

\begin{minipage}[t]{0.48\textwidth}
\centering
\includegraphics[width=\linewidth]{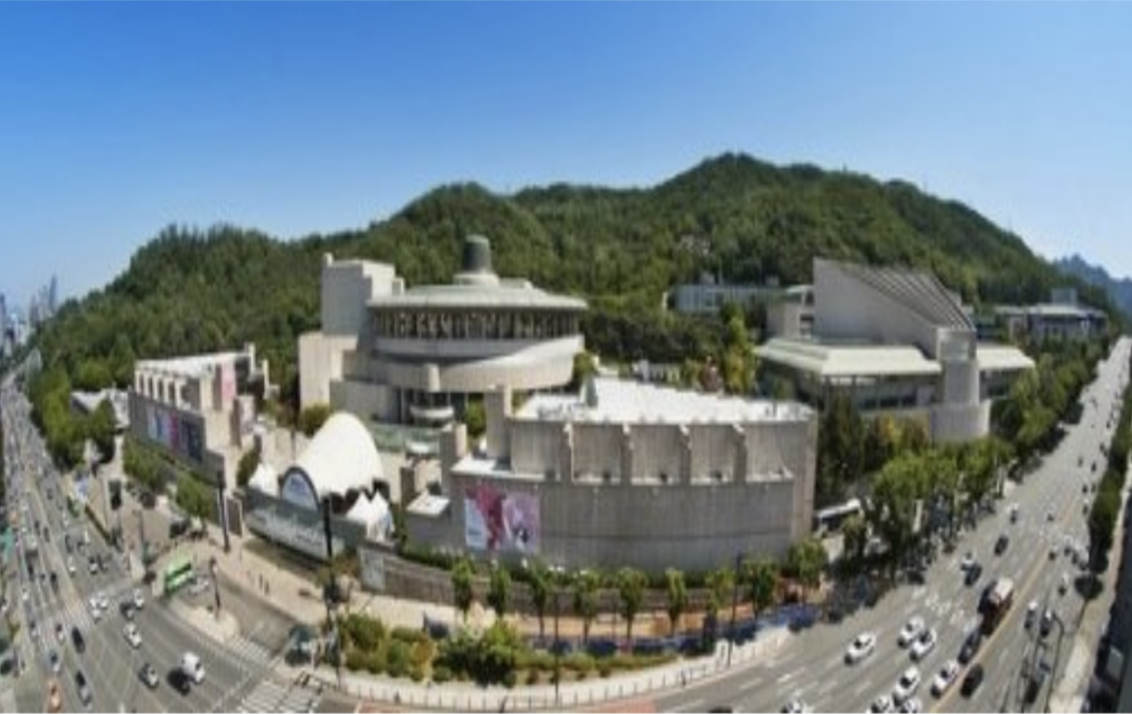} \\[0.5em]
\small [그림 1] 예술의전당 전경 \\
\small [Figure 1] Full View of Seoul Arts Center
\end{minipage}
\hfill
\begin{minipage}[t]{0.48\textwidth}
\centering
\includegraphics[width=\linewidth]{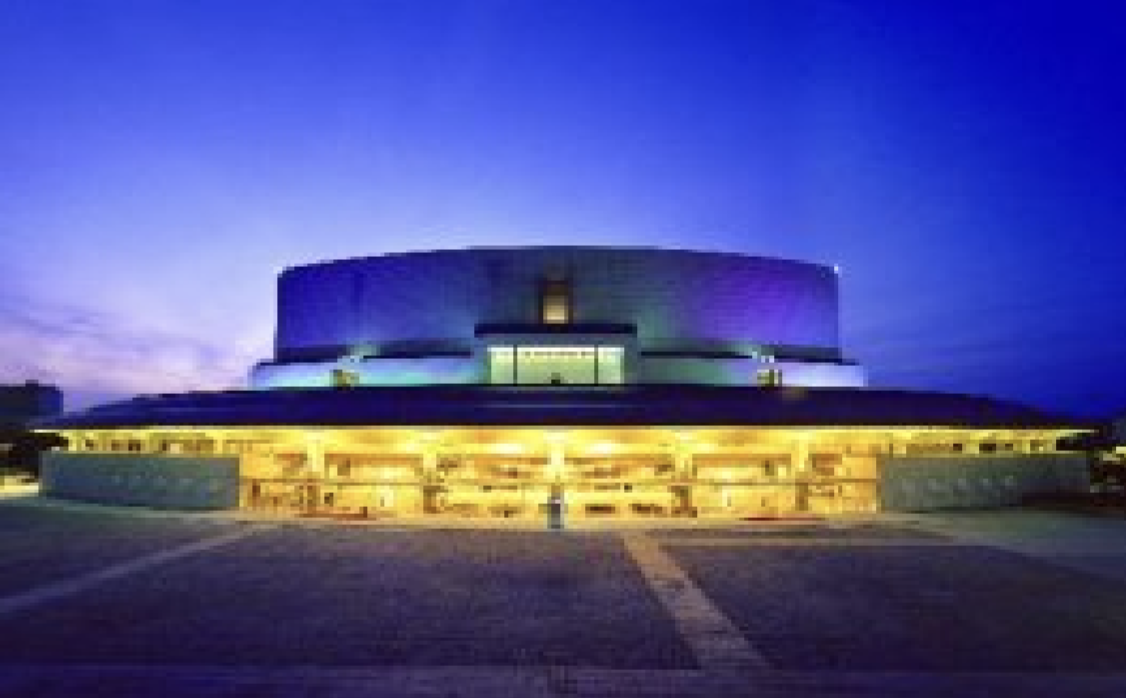} \\[0.5em]
\small [그림 2] 예술의전당 음악당 외관 \\
{\small [Figure 2] Exterior of SAC Music Hall}
\end{minipage}

\end{minipage}
\caption{Case 2 (Music) Introduction part of the PWYW questionnaire}
\label{fig:case2_intro}
\end{figure*}

\begin{figure*}[t]
\centering
\begin{minipage}{0.9\textwidth}
\small

\textbf{IV. 다음은 가상상황에 대한 설명입니다.} \\
\textbf{IV. The following is a hypothetical scenario.}

여러분은 현재 서울 서초구 예술의전당 근처에 있고 3–4시간 정도 여유시간이 있습니다.  
바로 옆 예술의전당에서는 성인기준 \textbf{B석 30,000원}으로 \textit{<손열음 피아노 독주회>}가 곧 시작할 예정입니다. 

You are currently near the Seoul Arts Center in Seocho-gu, Seoul, and have about 3–4 hours of free time.  
At the Arts Center nearby, the \textit{“Yiruma Son Piano Recital”} will begin soon, with \textbf{B-seat tickets priced at 30,000 KRW}.

\begin{center}
\includegraphics[width=0.2\textwidth]{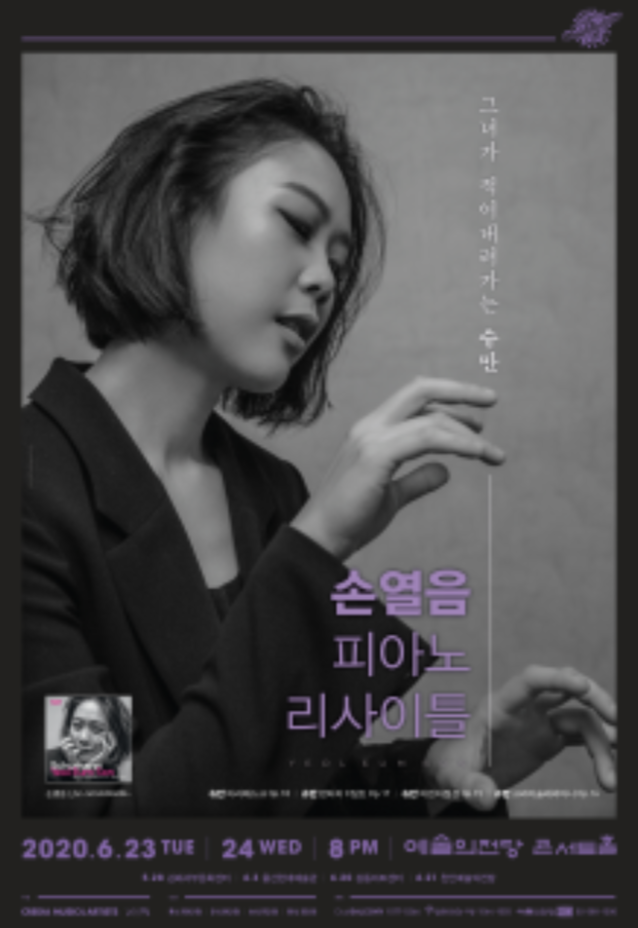} \\
{\small [그림] 공연 포스터 / [Figure] Performance Poster}
\end{center}

\textbf{<연주자 및 공연소개> About the Artist and Performance} \\
피아니스트 손열음은 2005년 루빈스타인 국제 피아노 콩쿠르에서 3위를 수상하고  
2009년 벨클라이번 국제 피아노 콩쿠르에서도 준우승한 대한민국 대표 피아니스트입니다. \\
In 2005, pianist Son Yeol-eum placed 3rd in the Rubinstein International Piano Competition,  
and in 2009 won the silver medal at the Van Cliburn International Piano Competition.  
She is one of Korea’s most celebrated classical pianists.

\vspace{0.5em}

이번 독주회는 그녀가 가장 아끼는 작곡가 슈만과 모차르트를 중심으로 구성된  
\textbf{“All Schumann Program”}으로, 손열음의 음악적 해석이 기대되는 무대입니다. \\
The recital will focus on her favorite composers — Schumann and Mozart — under the \textbf{“All Schumann Program”}.  
Audiences can look forward to her deep musical interpretation.

\vspace{1.5em}

\begin{enumerate}
\item 귀하께서는 \textbf{손열음 피아노 독주회를 B석 관람료 가격 30,000원을 지불하고 관람}할 의도가 있으십니까? \\
\textbf{Would you be willing to pay 30,000 KRW (B-seat) to attend the recital?} \\
① 아니요 (No) \hfill ② 그렇다 (Yes)

\vspace{1em}

\item 만일 \textbf{위 전시에 대해 방문객이 원하는 가격을 스스로 정해서} 지불할 수 있다면, 관람 의향이 있으십니까? \\
\textbf{If you could choose your own admission fee, would you still be willing to attend the recital?} \\
① 아니다 (No) $\rightarrow$ Q3로 이동 (Go to Q3) \\
② 그렇다 (Yes) $\rightarrow$ Q2-1로 이동 (Go to Q2-1)

\vspace{1em}

\item[Q2-1.] \textbf{방문하실 경우 얼마 정도 관람료로 낼 의향이} 있으십니까? \\
\textbf{If you choose to attend, how much would you be willing to pay for admission?} \\
1회 입장 시 \_\_\_\_ 원 (KRW)

\vspace{1em}

\item[Q3.] 귀하께서 \textbf{예술의전당 공연 관람 의사가 없다고 답변하신 이유는 무엇입니까?} \\
\textbf{If you are not willing to attend the performance, what are the reasons? (Check all that apply)} \\
\begin{itemize}
\item ① 해당 기관에서 제공하는 공연 및 프로그램에 관심이 없다 \\
\hspace{2em} I am not interested in the institution’s programs or performances.
\item ② 위 공연에 대한 프로그램이 만족스럽지 않다 \\
\hspace{2em} The program content is not satisfactory.
\item ③ 위 공연의 관람 여부를 판단하기 위한 정보가 충분히 주어지지 않았다 \\
\hspace{2em} There is insufficient information to make a decision.
\item ④ 나는 위의 공연과 관련된 주제에 관심이 없다 \\
\hspace{2em} I am not interested in the theme of this performance.
\item ⑤ 나는 상대적으로 다른 여가활동을 더 선호한다 \\
\hspace{2em} I prefer other leisure activities.
\item ⑥ 기타 (직접 서술): \_\_\_\_\_\_\_\_\_\_\_\_\_\_ \\
\hspace{2em} Other: \_\_\_\_\_\_\_\_\_\_\_\_\_\_
\end{itemize}
\end{enumerate}

\end{minipage}
\caption{Case 2 (Music) Question part of the PWYW questionnaire}
\label{fig:case2_question}
\end{figure*}

This section provides the full structure of the survey instrument used in the PWYW experiment. Figures from \ref{fig:survey_info} to \ref{fig:case2_question} illustrate the original Korean question forms shown to participants, including survey introductions and detailed question items for both art and music domains. All questions are presented in the original Korean followed by English translations in sequence. These materials are included for reference to clarify the experimental context and input provided to both human participants and LLMs.

\paragraph{Survey Introduction: }
See Figure \ref{fig:survey_info}

\paragraph{Art Survey Introduction:}
See Figure \ref{fig:case1_intro}

\paragraph{Art Survey Form:}
See Figure \ref{fig:case1_question}

\paragraph{Music Survey Introduction:}
See Figure \ref{fig:case2_intro}

\paragraph{Music Survey Form:}
See Figure \ref{fig:case2_question}

\section{Detailed Human Experiment Analysis}
To help readers understand, here we provide detailed result of the original human experiment, with 522 Korean adults. 

\subsection{Art Exhibition Analysis}
We employed the Heckman two-step model to analyze participants' Willingness-to-Pay (WTP) for art exhibitions, addressing potential sample selection bias inherent in Pay-What-You-Want (PWYW) scenarios.
\paragraph{Step 1: Selection Equation (Probit Model)}
In this step, we compute the inverse Mills ratio ($\lambda_i$), which can correct selection bias in the further steps.
Specifically, step 1 begins with assuming the probability of participation using a probit regression:

\begin{equation}
\Pr(\mathtt{A\_Q2} = 1) = \Phi(\gamma'Z),
\end{equation}
where the dependent variable \texttt{A\_Q2} is a binary indicator for willingness to participate in PWYW for art exhibition (1 = willing, 0 = not willing). And, $\Phi(\cdot)$ represents the cumulative distribution function (CDF) of the standard normal distribution.

For the analysis, we included the following independent variables in vector $Z$:
\begin{itemize}
\item \texttt{edu\_f}: Father's education level (1=middle school or below, 2=high school, 3=university, 4=graduate school) from \ref{app:persona.demographic}, Q8
\item \texttt{edu\_m}: Mother's education level (same coding) from \ref{app:persona.demographic}, Q9
\item \texttt{edu\_s}: Respondent's education level (same coding) from \ref{app:persona.demographic}, Q6
\item \texttt{art\_act\_dum}: Art activity participation dummy derived from \ref{app:persona.cult-part}, Q1 (1 if attended art exhibition $\geq 1$ time in past year, 0 otherwise)
\item \texttt{art\_att\_ave}: Average art attitude score, computed as mean of \ref{app:persona.cult-att}, Q7, Q10-Q15, Q19, and Q24 (5-point Likert scale attitudes toward art exhibitions, artists' professions, and cultural sophistication)
\item \texttt{art\_qz\_tot\_sco}: Art knowledge total score, sum of correct answers from \ref{app:persona.cult-know}, Q28-Q29 ("Mona Lisa" and "The Scream" painters, maximum score = 2)
\item \texttt{art\_ca\_edu\_b18}: Art cultural education before age 18, derived from \ref{app:persona.cult-ed}, Q2+Q4 (1 if received any art-related education in or outside school before age 18, 0 otherwise)
\item \texttt{art\_ca\_edu\_a19}: Art cultural education after age 19, derived from \ref{app:persona.cult-ed}, Q6+Q8 (1 if received any art-related education outside school after age 19, 0 otherwise)
\item \texttt{inc\_mo}: Monthly personal income (categorical: 1=none to 9=over 7 million KRW) from \ref{app:persona.demographic}, Q10
\item \texttt{marital\_dum}: Marital status dummy (1=married, 0=single/other) from \ref{app:persona.demographic}, Q3
\item \texttt{gen\_dum}: Gender dummy (0=male, 1=female) from \ref{app:persona.demographic}, Q1
\end{itemize}


Now, we calculate the inverse Mills ratio ($\lambda_i$) for each observation $Z_i$. The $\lambda_i$ serves as a selection correction term that accounts for potential bias arising from non-random sample selection in the PWYW decision-making process. We obtain $\lambda_i$ as follows:

\begin{equation}
\lambda_i = \frac{\mathcal{N}(\gamma'Z_i|0, 1^2)}{\Phi(\gamma'Z_i)},
\end{equation}
where $\mathcal{N}(\cdot|0, 1^2)$ represents the probability density function (PDF) of the standard normal distribution.\\

\paragraph{Step 2: Outcome Equation (Linear Regression with Selection Correction)}
Step 2 estimates willingness to pay amounts, corrected for selection bias using inverse Mills ratio $\lambda$:
\begin{equation}
\texttt{art\_pwyw\_log\_wtp} = \beta'X + \delta\lambda + \varepsilon,
\end{equation}
where the dependent variable \texttt{art\_pwyw\_log\_wtp} is the natural logarithm of WTP amount for art exhibition in KRW. And, $X$ is the vector representing independent variables with all $Z$ variables plus:
\begin{itemize}
\item \texttt{social\_n}: Social norm score from \ref{app:persona.social}, Q7 to Q10 (5-point Likert scale measuring social influence on behavior)
\item \texttt{moral\_n}: Moral norm score from \ref{app:persona.social}, Q2 to Q4 (5-point Likert scale measuring moral obligations and conscience)
\end{itemize}

\subsection{Music Performance Analysis}
The music analysis follows almost identical methodology where domain-specific variables changed to music-related variables, to analyze participants' WTP for music performances.
\paragraph{Step 1: Selection Equation (Probit Model)}
\begin{equation}
\Pr(\texttt{M\_Q2} = 1) = \Phi(\gamma'Z),
\end{equation}
where the dependent variable \texttt{M\_Q2} is a binary indicator for willingness to participate in PWYW for music performance. And, the $Z$ vector representing the following independent variables:
\begin{itemize}
\item \texttt{edu\_f}, \texttt{edu\_m}, \texttt{edu\_s}: Education variables (identical to art model)
\item \texttt{mus\_act\_dum}: Music activity participation dummy derived from \ref{app:persona.cult-part}, Q2 (1 if attended classical music/opera $\geq 1$ time in past year, 0 otherwise)
\item \texttt{mus\_att\_ave}: Average music attitude score, computed as mean of \ref{app:persona.cult-att}, Q7, Q10-Q18, Q20-Q23, and Q25 (5-point Likert scale attitudes toward classical music, opera, traditional music, and dance performances and professions, plus cultural sophistication)
\item \texttt{mus\_qz\_tot\_sco}: Music knowledge total score, sum of correct answers from \ref{app:persona.cult-know}, Q26-Q27 (Symphony "Fate" composer and "Rigoletto" composer, maximum score = 2)
\item \texttt{mus\_ca\_edu\_b18}: Music cultural education before age 18, derived \ref{app:persona.cult-ed}, Q1+Q3 (1 if received any music-related education in or outside school before age 18, 0 otherwise)
\item \texttt{mus\_ca\_edu\_a19}: Music cultural education after age 19, derived from \ref{app:persona.cult-ed}, Q5+Q7 (1 if received any music-related education outside school after age 19, 0 otherwise)
\item Sociodemographic controls: \texttt{inc\_mo}, \texttt{marital\_dum}, \texttt{gen\_dum} (identical to art model)
\end{itemize}

\paragraph{Step 2: Outcome Equation}
\begin{equation}
\texttt{mus\_pwyw\_log\_wtp} = \beta'X + \delta\lambda + \varepsilon,
\end{equation}
where the dependent variable \texttt{mus\_pwyw\_log\_wtp} is the natural logarithm of WTP amount for music performance. $X$ represents the set of independent variables including all variables in $Z$ plus:
\begin{itemize}
\item \texttt{social\_n\_ave}: Average social norm score from \ref{app:persona.social}, Q7 to Q10
\item \texttt{moral\_n\_ave}: Average moral norm score from \ref{app:persona.social}, Q2 to Q4
\end{itemize}

\subsection{Data Processing and Quality Control}
\paragraph{Sample Selection Criteria}
\begin{itemize}
\item Minimum 10 observations per group to ensure the convergence of statistical model
\item Minimum 2 categories in selection variable to satisfy variance requirements
\item Minimum 5 observations with positive WTP among those willing to participate
\end{itemize}
\paragraph{Variable Transformations}
\begin{itemize}
\item \textbf{WTP Amount}: We applied natural logarithm transformation to address skewness and heteroskedasticity.
\item \textbf{Cultural Education}: Binary coding by re-categorizing experiences into formal (school) and informal (outside school) education experiences.
\item \textbf{Attitude Scores}: Standardized averaging across relevant Likert scale items
\item \textbf{Knowledge Scores}: Simple summation of correctly answered responses
\end{itemize}

\paragraph{Extracted Summary of Statistical Model}
For each successful model estimation, we extracted:
\begin{itemize}
\item $\rho$: Correlation between error terms in selection and outcome equations
\item $\sigma$: Standard deviation of outcome equation error term
\item Wald $\chi^2$: Statistics from the test of overall model significance
\item McFadden $R^2$: Pseudo R-squared about the fitness of selection equation
\item Log likelihood: Indicator of fitness of statistical model
\item Sample composition: $N_{\textrm{selected}}$ and $N_{\textrm{nonselected}}$ observations
\item 
The statistical significance of the inverse Mills ratio ($\lambda$): Whether sample selection bias correction is necessary for valid inference
\end{itemize}

\section{Persona Input Formats}

Here, we illustrate how we provided the given persona as an input to LLMs.

\subsection{Survey Format}
In the Survey format condition, persona information was presented in a structured question-answer format in Korean, directly mirroring the original questionnaire structure used in the human experiment. Each persona was input as a systematic compilation of responses across all 65 survey items from the seven main categories described in Section A, maintaining the original Korean language to ensure linguistic consistency with the human experimental conditions.

Thus, the persona information was structured as follows:
\begin{verbatim}
What is your gender? : [Male/Female]
What is your age? : [Age in years]  
...
\end{verbatim}

\subsection{Storytelling Format}
In the Storytelling format condition, the same 65 survey items were transformed into cohesive biographical narratives using GPT-4o to create personalized life stories that naturally incorporated all persona attributes.


\subsection{Prompt for generating narratives}
The following prompt was used to generate storytelling format in Korean:

\textbf{Original Korean Prompt:}
\begin{quote}
아래 페르소나 정보를 기반으로, 한 여성 인물이 자신의 삶에서 예술을 어떻게 받아들이고 이해해 왔는지를 고백적으로 풀어내는 허구 수필을 작성하세요.
\begin{itemize}
    \item 텍스트에는 어떤 소제목, 부제, 섹션 제목도 포함되어서는 안 됩니다.
    \item 자연스러운 단락 구분은 허용되지만, 시각적 강조나 구획 나눔은 삼가세요.
    \item 전체 흐름은 시간 순이 아니어도 되며, 자유롭게 떠오르는 기억의 방식으로 구성해도 좋습니다.
    \item 지나치게 극적인 연출은 피하며 사실적인 정서에 기반해 구성하세요.
    \item 분량은 약 5000 토큰으로, 읽는 이로 하여금 한 사람의 내면을 들여다보는 느낌을 받을 수 있어야 합니다.
    \item \mbox{[PERSONA INFORMATION]}

\end{itemize}
\end{quote}

\textbf{English Translation:}
\begin{quote}
\textit{Based on the persona information below, write a confessional fictional essay that tells how a woman has embraced and understood art throughout her life.}
\begin{itemize}
    \item \textit{The text should not include any subtitles, subtitles, or section titles.}
    \item \textit{Natural paragraph breaks are allowed, but visual emphasis or partitioning should be avoided.}
    \item \textit{The overall flow does not have to be chronological, and can be freely composed in the way of emerging memories.}
    \item \textit{Avoid overly dramatic production and compose based on realistic emotions.}
    \item \textit{The volume should be about 5000 tokens, so that the reader should feel like looking into the inner side of a person.}
    \item \mbox{[PERSONA INFORMATION]}
\end{itemize}
\end{quote}

We used the following environment to generate narratives.
\begin{itemize}
    \item \textbf{API}: OpenAI official API
    \item \textbf{Model}: GPT-4o
    \item \textbf{Temperature}: 0
    \item \textbf{Max tokens}: 6,000
\end{itemize}

\begin{figure*}[t]  
\centering
\begin{minipage}{0.95\textwidth}
\small

\textbf{국·공립 박물관 및 공연장 관람료에 대한 소비자 설문조사} \\
\textbf{Consumer Survey on Admission Fees for National/Public Museums and Performance Halls}

\vspace{1em}

\textbf{[대상 안내 / Participant Criteria]}\\
본 설문은 \textbf{30대 (만 30세에서 만 39세)}를 대상으로 진행됩니다.  
\textbf{국립중앙박물관 특별전 「시대의 얼굴, 셰익스피어에서 에드 시런까지 2021」을 관람하지 않은 분},  
\textbf{손열음 피아노 독주회를 관람하지 않은 분}만 응답 가능합니다.  
위 전시 또는 공연을 관람하신 분은 설문을 종료해 주세요.

This survey targets individuals in their 30s (ages 30–39).  
It is intended only for those who have \textbf{not visited the special exhibition "The Face of the Era"} at the National Museum of Korea  
and have \textbf{not attended the Sunwook Kim piano recital}.  
If you have seen either, please discontinue this survey.

\vspace{1em}

\textbf{[조사 목적 / Survey Purpose]}\\
본 설문은 국·공립 박물관 및 공연장 관람료 지불의사에 대한 연구 목적을 가지고 있으며,  
응답 내용은 설문 목적 외의 용도로 절대 사용되지 않습니다.  
귀하의 응답은 문화예술산업의 발전을 위한 소중한 자료로 활용될 예정입니다.  
바쁘시더라도 성실한 응답을 부탁드립니다.

This survey investigates willingness to pay for admissions to national/public museums and performance halls.  
All answers will be used solely for research purposes and will never be used beyond the scope of the survey.  
Your responses are valuable for advancing the cultural and arts industry. We kindly ask for your honest participation.

\vspace{1em}

\textbf{[응답자 보호 / Confidentiality]}\\
수집된 정보는 \textit{통계법} 제33조(비밀의 보호), 제34조(조사자의 의무)에 따라 보호되며,  
귀하의 정보는 익명으로 처리됩니다. 어떠한 경우에도 신상 정보가 노출되지 않습니다.

In accordance with Articles 33 and 34 of the Statistics Act, all collected information will be protected and kept strictly confidential.  
Your answers will remain anonymous, and no identifying information will be disclosed under any circumstances.

\vspace{1em}

※ 각 문항에는 옳고 그른 답이 없으며, 솔직하게 평소 생각을 바탕으로 응답해 주시기 바랍니다.\\
※ There are no right or wrong answers. Please respond honestly based on your usual thoughts.

\vspace{1em}

다시 한 번, 설문에 참여해 주셔서 감사합니다.\\
Once again, thank you very much for your participation.

\end{minipage}
\caption{Survey Introduction part of the PWYW questionnaire}
\label{fig:survey_info}
\end{figure*}

\section{Detailed Result for RQ1}
This section presents the regression outcomes of the RQ1 experimental conditions using the Heckman two-step model. Each table pair (Tables from \ref{tab:structure_art} to \ref{tab:human_guide_music}) summarizes how LLMs (GPT-4o, LLaMA, Qwen) reflect persona attributes when predicting willingness-to-pay decisions in the PWYW task. The tables are provided for reference to illustrate model-level behavioral patterns under different interaction settings.

\paragraph{Sequential Condition}  
See Tables \ref{tab:structure_art} and \ref{tab:structure_music}.

\begin{table*}[]
    \centering
    \footnotesize
    \begin{tabular}{l|c@{\;\;}c@{\;\;}c@{\;\;}r|c@{\;\;}c@{\;\;}c@{\;\;}r|c@{\;\;}c@{\;\;}c@{\;\;}r}
    \toprule
    & \multicolumn{4}{c}{GPT-4o} & \multicolumn{4}{c}{LLaMA} & \multicolumn{4}{c}{Qwen} \\
         & coeff & std\_error& p value & z value & coeff & std\_error & p value &  z value  & coeff & std\_error & p value & z value \\
    \midrule
    \multicolumn{10}{l}{\textbf{Outcome}} \\
    edu\_f & 0.01 & 0.01 & 0.21   & 1.26      & 0.04 & 21694.32 & 1.00 & 0.00    & 0.01 & 0.02 & 0.51 & 0.66 \\
    edu\_m & -0.01 & 0.01  & 0.12 & -1.57    & -0.03 & 25017.10 & 1.00 & 0.00     & 0.00 & 0.02 & 0.90 & 0.13 \\
    mus\_act\_dum & 0.02 & 0.01 & p$<$0.05  & 2.00    & -0.02 & 303822.33 & 1.00 & 0.00      & 0.06 & 0.03 & p$<$0.05 & 2.38 \\
    mus\_att\_ave & 0.02 & 0.02 & 0.24  & 1.18     & 0.19 & 22875.96 & 1.00 & 0.00     & 0.26  & 0.03 & p$<$0.0001 & 0.02 \\
    mus\_ca\_edu\_a19 & 0.00 & 0.01 & 0.52 & 0.64    & -0.03 & 24289.04 & 1.00 & 0.00     & 0.03 & 0.02 & 0.19 & 10.00 \\
    mus\_ca\_edu\_b18 & 0.00 & 0.01 & 0.64 & 0.47    & 0.06 & 18689.85 & 1.00 & 0.00     & 0.00 & 0.02 & 0.86 & 1.31 \\
    mus\_qz\_tot\_sco & 0.01 & 0.01 & 0.42 & 0.81    & 0.12 & 25466.45 & 1.00 & 0.00     & 0.05 & 0.02 & p$<$0.05 & 0.18 \\
    moral\_n\_ave & 0.00 & 0.01 & 0.90 & 0.12     & 0.04 & 21456.43 & 1.00 & 0.00     & 0.04 & 0.02 & p$<$0.05 & 2.12 \\
    social\_n\_ave & 0.00 & 0.01 & 0.65 & -0.46    & 0.04 & 19449.28 & 1.00 & 0.00     & 0.00 & 0.02 & 0.94 & 0.08 \\
    inc\_mo & 0.00 & 0.00 & 0.87 & -0.16         & 0.03 & 7365.53  & 1.00 & 0.00     & 0.00 & 0.01 & 0.65 & 0.45 \\
    matarial\_dum & 0.00 & 0.01 & 0.71 & 0.37    & 0.08 & 27206.99 & 1.00 & 0.00     & 0.04 & 0.02 & 0.66 & 1.88 \\
    \textit{constant} & 8.98 & 0.08 & p$<$0.001 & 106.26      & 7.33 & 103897.22 & 1.00 & 0.00     & 7.28 & 0.12 & p$<$0.0001 & 61.51 \\
    \midrule
    \multicolumn{10}{l}{\textbf{Model Statistics}} \\
    Wald $\chi^2$ & \multicolumn{4}{c}{9.59} & \multicolumn{4}{c}{3.1$\times 10^{-10}$}& \multicolumn{4}{c}{183.19} \\
    $\rho$ & \multicolumn{4}{c}{0.69} & \multicolumn{4}{c}{1}& \multicolumn{4}{c}{0.73} \\
    $\sigma$ & \multicolumn{4}{c}{0.07} & \multicolumn{4}{c}{299218.13}& \multicolumn{4}{c}{0.25} \\
    Log Likelihood & \multicolumn{4}{c}{-201.81} & \multicolumn{4}{c}{-}& \multicolumn{4}{c}{-83.74} \\
    McFadden $R^2$ & \multicolumn{4}{c}{0.44} & \multicolumn{4}{c}{-}& \multicolumn{4}{c}{0.59} \\
    $N_\textrm{nonselected}$ & \multicolumn{4}{c}{268} & \multicolumn{4}{c}{1}& \multicolumn{4}{c}{69} \\
    $N_\textrm{selected}$ & \multicolumn{4}{c}{254} & \multicolumn{4}{c}{521}& \multicolumn{4}{c}{453} \\
    \bottomrule
    \end{tabular}
    \caption{Regression outcomes of Structure Condition on the art PWYW decision-making task. The table reports coefficients, standard errors, z-values, and p-values from a Heckman two-step model. Bottom panel presents model-level statistics including selection bias correction and sample distributions.}
    \label{tab:structure_art}
\end{table*}

\begin{table*}[]
    \centering
    \footnotesize
    \begin{tabular}{l|c@{\;\;}c@{\;\;}c@{\;\;}c|c@{\;\;}c@{\;\;}c@{\;\;}c|c@{\;\;}c@{\;\;}c@{\;\;}c}

    \toprule
    & \multicolumn{4}{c}{GPT-4o} & \multicolumn{4}{c}{LLaMA} & \multicolumn{4}{c}{Qwen} \\
         & coeff & std\_error& p value & z value & coeff & std\_error & p value &  z value  & coeff & std\_error & p value & z value \\
    \midrule
    \multicolumn{10}{l}{\textbf{Outcome}} \\
    edu\_f & -0.01 & 0.02 & 0.64 & -0.47     & 0.04 & 21694.32 & 1.00 & 0.00     & 0.01 & 0.02 & 0.51 & 0.66\\
    edu\_m & 0.00 & 0.03 & 0.10 & 0.01     & -0.03 & 25017.10 & 1.00 & 0.00     & 0.00 & 0.02 & 0.90 & 0.13\\
    mus\_act\_dum & 0.05 & 0.04 & 0.19 & 1.30     & -0.02 & 30822.33 & 1.00 & 0.00     & 0.06 & 0.03 & p$<$0.05 & 2.38\\
    mus\_att\_ave & 0.19 & 0.06 & p$<$0.01 & 3.17     & 0.19 & 22875.96 & 1.00 & 0.00     & 0.26 & 0.03 & p$<$0.0001 & 10.00\\
    mus\_ca\_edu\_a19 & 0.00 & 0.02 & 0.92 & -0.11     & -0.03 & 24289.04 & 1.00 & 0.00     & 0.03 & 0.02 &0.19 & 1.31\\
    mus\_ca\_edu\_b18 & 0.03 & 0.02 & 0.16 & -1.39     & 0.06 & 18689.85 & 1.00 & 0.00     & 0.00 & 0.02 & 0.86 & 0.18\\
    mus\_qz\_tot\_sco & 0.00 & 0.02 & 0.85 & -0.19     & 0.12 & 25466.48 & 1.00 & 0.00     & 0.05 & 0.02 & p$<$0.05 & 2.15\\
    moral\_n\_ave & 0.04 & 0.02 & 0.09 & 1.67     & 0.04 & 21456.43 & 1.00 & 0.00     & 0.04 & 0.02 & p$<$0.05 & 2.12\\
    social\_n\_ave & 0.01 & 0.02 & 0.08 & 0.51     & 0.04 & 19449.28 & 1.00 & 0.00     & 0.00 & 0.02 & 0.94 & 0.08\\
    inc\_mo & 0.01 & 0.01 & 0.71  & 1.74     & 0.03 & 7365.53 & 1.00 & 0.00     & 0.00 & 0.01 & 0.65 & 0.45\\
    matarial\_dum & 0.01 & 0.03 & p$<$0.05 & 0.36     & 0.08 & 27206.99 & 1.00 & 0.00     & 0.04 & 0.02 & 0.06 & 1.88\\
    \textit{constant} & 8.87 & 0.24 & p$<$0.0001 & 36.91     & 7.33 & 103897.22 & 1.00 & 0.00     & 7.28 & 0.12 & p$<$0.0001 & 61.51\\
    \midrule
    \multicolumn{10}{l}{\textbf{Model Statistics}} \\
    Wald $\chi^2$ & \multicolumn{4}{c}{9.59} & \multicolumn{4}{c}{3.1$\times 10^{-10}$}& \multicolumn{4}{c}{183.19} \\
    $\rho$ & \multicolumn{4}{c}{0.69} & \multicolumn{4}{c}{1}& \multicolumn{4}{c}{0.73} \\
    $\sigma$ & \multicolumn{4}{c}{0.07} & \multicolumn{4}{c}{299218.13}& \multicolumn{4}{c}{0.25} \\
    Log Likelihood & \multicolumn{4}{c}{-201.81} & \multicolumn{4}{c}{-}& \multicolumn{4}{c}{-83.74} \\
    McFadden $R^2$ & \multicolumn{4}{c}{0.44} & \multicolumn{4}{c}{-}& \multicolumn{4}{c}{0.59} \\
    $N_\textrm{nonselected}$ & \multicolumn{4}{c}{268} & \multicolumn{4}{c}{1}& \multicolumn{4}{c}{69} \\
    $N_\textrm{selected}$ & \multicolumn{4}{c}{254} & \multicolumn{4}{c}{521}& \multicolumn{4}{c}{453} \\
    \bottomrule
    \end{tabular}
    \caption{Regression outcomes of Structure Condition on the music PWYW decision-making task. The table reports coefficients, standard errors, z-values, and p-values from a Heckman two-step model. Bottom panel presents model-level statistics including selection bias correction and sample distributions.}
    \label{tab:structure_music}
\end{table*}

\paragraph{Human-Guided Condition}  
See Tables \ref{tab:human_guide_art} and \ref{tab:human_guide_music}. 
\begin{table*}[]
    \centering
    \footnotesize
    \begin{tabular}{l|c@{\;\;}c@{\;\;}c@{\;\;}c|c@{\;\;}c@{\;\;}c@{\;\;}c|c@{\;\;}c@{\;\;}c@{\;\;}c}

    \toprule
    & \multicolumn{4}{c}{GPT-4o} & \multicolumn{4}{c}{LLaMA} & \multicolumn{4}{c}{Qwen} \\
         & coeff & std\_error& p value & z value & coeff & std\_error & p value &  z value  & coeff & std\_error & p value & z value \\
    \midrule
    \multicolumn{10}{l}{\textbf{Outcome}} \\
    edu\_f & -0.02 & 0.05 & 0.73 & -0.34     & 0.03 & 0.08 & 0.71 & 0.37     & -0.09 & 0.17 & 0.58 & -0.55 \\
    edu\_m & 0.04 & 0.06 & 0.47 & 0.72     & -0.02 & 0.09 & 0.82 & -0.22      & 0.12 & 0.20 & 0.53 & 0.62 \\
    mus\_act\_dum & -0.04 & 0.12 & 0.73 & -0.34     & 0.14 & 0.19 & 0.47 & -0.72     & -0.25 & 0.41 & 0.54 & -0.62 \\
    mus\_att\_ave & 0.07 & 0.11 & 0.48 & 0.70     & 0.11 & 0.17 & 0.55 & 0.60     & -0.25 & 0.37 & 0.50 & -0.67 \\
    mus\_ca\_edu\_a19 & 0.01 & 0.04 & 0.75 & 0.32     & 0.04 & 0.05 & 0.50 & 0.68     & -0.06 & 0.13 & 0.63 & -0.48 \\
    mus\_ca\_edu\_b18 & -0.04 & 0.05 & 0.40 & -0.84     & -0.03 & 0.08 & 0.76 & -0.31     & -0.12 & 0.18 & 0.50 & -0.68 \\
    mus\_qz\_tot\_sco & -0.02 & 0.05 & 0.67 & -0.43     & 0.03 & 0.08 & 0.72 & 0.35     & -0.08 & 0.17 & 0.62 & -0.50 \\
    moral\_n\_ave & 0.04 & 0.03 & 0.19 & 1.30     & 0.10 & 0.05 & p$<$0.05 & 2.07     & -0.02 & 0.10 & 0.87 & -0.17 \\
    social\_n\_ave & 0.00 & 0.03 & 0.93 & 0.08     & 0.03 & 0.05 & 0.49 & 0.69     & 0.07 & 0.09  & 0.46 & 0.74 \\
    inc\_mo & 0.01 & 0.01 & 0.28 & 1.09     & 0.03 & 0.07 & 0.05 & 1.93     & -0.01 & 0.04  & 0.76 & -0.31  \\
    matarial\_dum & 0.01 & 0.05 & 0.90 & 0.13     & 0.00 & 0.92 & 0.94 & 0.07     & -0.03 & 0.16  & 0.83 & -0.22 \\
    \textit{constant} & 8.50 & 0.56 & p$<$0.0001 & 15.28     & 7.68 & 1.65 & p$<$0.0001 & 8.32     & 9.91 & 1.95 & p$<$0.0001 & 5.08 \\
    \midrule
    \multicolumn{10}{l}{\textbf{Model Statistics}} \\
    Wald $\chi^2$ & \multicolumn{4}{c}{20.11} & \multicolumn{4}{c}{17.88}& \multicolumn{4}{c}{2.04} \\
    $\rho$ & \multicolumn{4}{c}{-1} & \multicolumn{4}{c}{0.02}& \multicolumn{4}{c}{-1} \\
    $\sigma$ & \multicolumn{4}{c}{0.46} & \multicolumn{4}{c}{0.62}& \multicolumn{4}{c}{1.60} \\
    Log Likelihood & \multicolumn{4}{c}{-250.78} & \multicolumn{4}{c}{-250.78}& \multicolumn{4}{c}{-250.78} \\
    McFadden $R^2$ & \multicolumn{4}{c}{0.08} & \multicolumn{4}{c}{0.08}& \multicolumn{4}{c}{0.08} \\
    $N_\textrm{nonselected}$ & \multicolumn{4}{c}{113} & \multicolumn{4}{c}{113}& \multicolumn{4}{c}{113} \\
    $N_\textrm{selected}$ & \multicolumn{4}{c}{409} & \multicolumn{4}{c}{409}& \multicolumn{4}{c}{409} \\
    \bottomrule
    \end{tabular}
    \caption{Regression outcomes of Human-Guided Condition on the art PWYW decision-making task. The table reports coefficients, standard errors, z-values, and p-values from a Heckman two-step model. Bottom panel presents model-level statistics including selection bias correction and sample distributions.}
    \label{tab:human_guide_art}
\end{table*}

\begin{table*}[]
    \centering
    \footnotesize
    \begin{tabular}{l|c@{\;\;}c@{\;\;}c@{\;\;}c|c@{\;\;}c@{\;\;}c@{\;\;}c|c@{\;\;}c@{\;\;}c@{\;\;}c}

    \toprule
    & \multicolumn{4}{c}{GPT-4o} & \multicolumn{4}{c}{LLaMA} & \multicolumn{4}{c}{Qwen} \\
        & coeff & std\_error& p value & z value & coeff & std\_error & p value &  z value  & coeff & std\_error & p value & z value \\
    \midrule
    \multicolumn{10}{l}{\textbf{Outcome}} \\
    edu\_f & -0.09 & 0.05 & 0.05 & -1.95      & -0.05 & 0.06 & 0.41 & -0.82     & -0.04 & 0.03 & 0.20 & -1.29 \\
    edu\_m & -0.03 & 0.05 & 0.54 & 0.61     & -0.02 & 0.05 & 0.65 & -0.46     & -0.02 & 0.03 & 0.63 & -0.48 \\
    mus\_act\_dum & -0.09 & 0.11 & 0.39 & -0.85     & -0.02 & 0.12 & 0.89 & -0.13     & -0.03 & 0.07 & 0.70 & -0.38 \\
    mus\_att\_ave & -0.08 & 0.14 & 0.57 & -0.57     & -0.10 & 0.17 & 0.57 & -0.57     & 0.00 & 0.10 & 0.98 & -0.02 \\
    mus\_ca\_edu\_a19 & 0.02 & 0.04 & 0.69 & 0.40     & 0.04 & 0.05 & 0.38 & 0.88     & 0.03 & 0.03 & 0.38 & 0.87 \\
    mus\_ca\_edu\_b18 & 0.00 & 0.04 & 0.91 & 0.11     & -0.02 & 0.05 & 0.61 & -0.51     & -0.02 & 0.03 & 0.51 & -0.66  \\
    mus\_qz\_tot\_sco & 0.05 & 0.05 & 0.30 & 1.04     & 0.11 & 0.06 & p$<$0.05 & 2.03     & 0.06 & 0.03 & 0.07 & 1.84 \\
    moral\_n\_ave & 0.13 & 0.03 & p$<$0.0001 & 3.95     & 0.13 & 0.04 & p$<$0.0001 & 2.99     & 0.02 & 0.02 & 0.37 & -0.90 \\
    social\_n\_ave & -0.01 & 0.03 & 0.73 & -0.34     & -0.02 & 0.04 & 0.70 & -0.39     & 0.02 & 0.02 & 0.38 & 0.88 \\
    inc\_mo & 0.03 & 0.01 & p$<$0.05 & 2.27     & -0.03 & 0.02 & 0.09 & 1.70     & 0.01 & 0.01 & 0.44 & 0.78 \\
    matarial\_dum & -0.04 & 0.05 & 0.46 & -0.73     & -0.04 & 0.06 & 0.46 & -0.75     & 0.01 & 0.04 & 0.82 & 0.23 \\
    \textit{constant} & 9.61 & 0.59 & p$<$0.0001 & 16.24     & 9.75 & 0.72 & p$<$0.0001 & 13.52     & 9.77 & 0.42 & p$<$0.0001 & 23.22 \\
    \midrule
    \multicolumn{10}{l}{\textbf{Model Statistics}} \\
    Wald $\chi^2$ & \multicolumn{4}{c}{39.04} & \multicolumn{4}{c}{31.44}& \multicolumn{4}{c}{20.32} \\
    $\rho$ & \multicolumn{4}{c}{-0.98} & \multicolumn{4}{c}{-0.54}& \multicolumn{4}{c}{-0.87} \\
    $\sigma$ & \multicolumn{4}{c}{0.49} & \multicolumn{4}{c}{0.53}& \multicolumn{4}{c}{0.34} \\
    Log Likelihood & \multicolumn{4}{c}{-290.01} & \multicolumn{4}{c}{290.01}& \multicolumn{4}{c}{-290.01} \\
    McFadden $R^2$ & \multicolumn{4}{c}{0.16} & \multicolumn{4}{c}{0.16}& \multicolumn{4}{c}{0.16} \\
    $N_\textrm{nonselected}$ & \multicolumn{4}{c}{195} & \multicolumn{4}{c}{195}& \multicolumn{4}{c}{195} \\
    $N_\textrm{selected}$ & \multicolumn{4}{c}{327} & \multicolumn{4}{c}{327}& \multicolumn{4}{c}{327} \\
    \bottomrule
    \end{tabular}
    \caption{Regression outcomes of Human-Guided Condition on the music PWYW decision-making task. The table reports coefficients, standard errors, z-values, and p-values from a Heckman two-step model. Bottom panel presents model-level statistics including selection bias correction and sample distributions.}
    \label{tab:human_guide_music}
\end{table*}

\section{Detailed Result for RQ2}
This section presents the regression outcomes of various RQ2 experimental conditions using the Heckman two-step model. Each table pair (Tables from \ref{tab:survey_cot_art} to \ref{tab:story_rag_music}) summarizes how LLMs (GPT-4o, LLaMA, Qwen) reflect persona attributes under different combinations of persona format (Survey vs. Storytelling) and prompting method (Base, CoT, RAG, Few-shot). The tables are provided for reference to illustrate model-level behavioral patterns.

\paragraph{Survey Format - CoT}
See Tables \ref{tab:survey_cot_art} and \ref{tab:survey_cot_music} 
\begin{table*}[]
    \centering
    \footnotesize
    \begin{tabular}{l|c@{\;\;}c@{\;\;}c@{\;\;}c|c@{\;\;}c@{\;\;}c@{\;\;}c|c@{\;\;}c@{\;\;}c@{\;\;}c}

    \toprule
    & \multicolumn{4}{c}{GPT-4o} & \multicolumn{4}{c}{LLaMA} & \multicolumn{4}{c}{Qwen} \\
        & coeff & std\_error& p value & z value & coeff & std\_error & p value &  z value  & coeff & std\_error & p value & z value \\
    \midrule
    \multicolumn{10}{l}{\textbf{Outcome}} \\
    edu\_f & 0.00 & 0.05 & 0.99 & -0.01      & -0.01 & 0.02 & 0.60 & -0.52      & 0.03 & 0.03 & 0.31 & 1.01\\
    edu\_m & 0.01 & 0.06 & 0.92 & 0.10      & 0.04 & 0.03 & 0.15 & 1.44      & -0.01 & 0.03 & 0.84 & -0.20 \\
    mus\_act\_dum & 0.08 & 0.13 & 0.51 & 0.66      & 0.00 & 0.05 & 0.95 & 0.06      & 0.08 & 0.06 & 0.19 & 1.32\\
    mus\_att\_ave & 0.21 & 0.11 & 0.07 & 1.81      & 0.05 & 0.05 & 0.29 & 1.05      & 0.14 & 0.06 & p$<$0.05 & 2.47\\
    mus\_ca\_edu\_a19 & 0.00 & 0.04 & 0.97 & -0.04      & 0.00 & 0.02 & 0.80 & 0.25      & 0.01 & 0.02 & 0.75 & 0.32\\
    mus\_ca\_edu\_b18 & 0.03 & 0.05 & 0.56 & 0.58      & -0.01 & 0.02 & 0.75 & -0.32      & -0.01 & 0.03 & 0.74 & -0.33\\
    mus\_qz\_tot\_sco & 0.04 & 0.05 & 0.48 & 0.71      & -0.02 & 0.02 & 0.32 & -0.99      & 0.01 & 0.03 & 0.72 & 0.36\\
    moral\_n\_ave & 0.04 & 0.03 & 0.16 & 1.42      & 0.03 & 0.01 & 0.07 & 1.79      & 0.01 & 0.02 & 0.43 & 0.80\\
    social\_n\_ave & -0.01 & 0.03 & 0.79 & -0.27      & 0.00 & 0.01 & 0.80 & 0.25      & 0.00 & 0.01 & 0.78 & -0.28\\
    inc\_mo & 0.01 & 0.01 & 0.48 & 0.70      & 0.00 & 0.01 & 0.38 & 0.88      & 0.00 & 0.01 & 0.79 & -0.27\\
    matarial\_dum & 0.04 & 0.05 & 0.39 & 0.86      & 0.01 & 0.02 & 0.58 & 0.55      & 0.01 & 0.02 & 0.53 & 0.62\\
    \textit{constant} & 7.71 & 0.60 & p$<$0.0001 & 12.85      & 8.62 & 0.27 & p$<$0.0001 & 32.51      & 8.11 & 0.29 & p$<$0.0001 & 27.55\\
    \midrule
    \multicolumn{10}{l}{\textbf{Model Statistics}} \\
    Wald $\chi^2$ & \multicolumn{4}{c}{16.42} & \multicolumn{4}{c}{37.77}& \multicolumn{4}{c}{26.87} \\
    $\rho$ & \multicolumn{4}{c}{1.00} & \multicolumn{4}{c}{-0.43}& \multicolumn{4}{c}{0.86} \\
    $\sigma$ & \multicolumn{4}{c}{0.49} & \multicolumn{4}{c}{0.18}& \multicolumn{4}{c}{0.23} \\
    Log Likelihood & \multicolumn{4}{c}{-250.78} & \multicolumn{4}{c}{-250.78}& \multicolumn{4}{c}{-250.78} \\
    McFadden $R^2$ & \multicolumn{4}{c}{0.08} & \multicolumn{4}{c}{0.08}& \multicolumn{4}{c}{0.08} \\
    $N_\textrm{nonselected}$ & \multicolumn{4}{c}{113} & \multicolumn{4}{c}{113}& \multicolumn{4}{c}{113} \\
    $N_\textrm{selected}$ & \multicolumn{4}{c}{409} & \multicolumn{4}{c}{409}& \multicolumn{4}{c}{409} \\
    \bottomrule
    \end{tabular}
    \caption{Regression outcomes of Survey\_CoT on the art PWYW decision-making task. The table reports coefficients, standard errors, z-values, and p-values from a Heckman two-step model. Bottom panel presents model-level statistics including selection bias correction and sample distributions.}
    \label{tab:survey_cot_art}
\end{table*}

\begin{table*}[]
    \centering
    \footnotesize
   \begin{tabular}{l|c@{\;\;}c@{\;\;}c@{\;\;}c|c@{\;\;}c@{\;\;}c@{\;\;}c|c@{\;\;}c@{\;\;}c@{\;\;}c}

    \toprule
    & \multicolumn{4}{c}{GPT-4o} & \multicolumn{4}{c}{LLaMA} & \multicolumn{4}{c}{Qwen} \\
         & coeff & std\_error& p value & z value & coeff & std\_error & p value &  z value  & coeff & std\_error & p value & z value \\
    \midrule
    \multicolumn{10}{l}{\textbf{Outcome}} \\
    edu\_f & -0.10 & 0.07 & 0.13 & -1.50      & -0.03 & 0.06 & 0.53 & -0.63      & -0.06 & 0.05 & 0.28 & -1.08\\
    edu\_m & 0.03 & 0.07 & 0.62 & 0.49      & -0.02 & 0.05 & 0.64 & -0.46      & 0.00 & 0.05 & 0.93 & -0.08 \\
    mus\_act\_dum & -0.16 & 0.15 & 0.29 & -1.07      & -0.02 & 0.12 & 0.89 & -0.14      & -0.03 & 0.12 & 0.81 & -0.24\\
    mus\_att\_ave & -0.07 & 0.20 & 0.71 & -0.37      & -0.12 & 0.17 & 0.50 & -0.68      & -0.16 & 0.16 & 0.34 & -0.96\\
    mus\_ca\_edu\_a19 & 0.03 & 0.06 & 0.60 & 0.52      & 0.03 & 0.05 & 0.50 & 0.68      & 0.03 & 0.05 & 0.55 & 0.61\\
    mus\_ca\_edu\_b18 & -0.04 & 0.06 & 0.53 & -0.62      & -0.02 & 0.05 & 0.63 & -0.48      & -0.03 & 0.05 & 0.52 & -0.64\\
    mus\_qz\_tot\_sco & 0.04 & 0.07 & 0.53 & 0.63      & 0.11 & 0.06 & 0.05 & 1.95      & 0.11 & 0.06 & p$<$0.05 & 2.05\\
    moral\_n\_ave & 0.07 & 0.05 & 0.10 & 1.63      & 0.13 & 0.04 & p$<$0.0001 & 3.12      & 0.09 & 0.04 & p$<$0.05 & 2.20\\
    social\_n\_ave & 0.02 & 0.04 & 0.59 & 0.54      & -0.02 & 0.04 & 0.71 & -0.38      & 0.04 & 0.04 & 0.32 & 0.98\\
    inc\_mo & 0.02 & 0.02 & 0.31 & 1.01      & 0.02 & 0.02 & 0.14 & 1.48      & 0.02 & 0.02 & 0.29 & 1.06\\
    matarial\_dum & 0.01 & 0.07 & 0.92 & 0.10      & -0.07 & 0.06 & 0.27 & -1.11      & -0.03 & 0.06 & 0.58 & -0.56\\
    \textit{constant} & 9.69 & 0.83 & p$<$0.0001 & 11.68      & 9.80 & 0.72 & p$<$0.0001 & 13.65      & 10.00 & 0.69 & p$<$0.0001 & 14.43\\
    \midrule
    \multicolumn{10}{l}{\textbf{Model Statistics}} \\
    Wald $\chi^2$ & \multicolumn{4}{c}{14.54} & \multicolumn{4}{c}{29.98} & \multicolumn{4}{c}{23.01} \\
    $\rho$ & \multicolumn{4}{c}{-1.00} & \multicolumn{4}{c}{-0.58} & \multicolumn{4}{c}{-0.90} \\
    $\sigma$ & \multicolumn{4}{c}{0.70} & \multicolumn{4}{c}{0.53} & \multicolumn{4}{c}{0.56}\\
    Log Likelihood & \multicolumn{4}{c}{-290.01} & \multicolumn{4}{c}{-290.01} & \multicolumn{4}{c}{-290.01}\\
    McFadden $R^2$ & \multicolumn{4}{c}{0.16} & \multicolumn{4}{c}{0.16} & \multicolumn{4}{c}{0.16} \\
    $N_\textrm{nonselected}$ & \multicolumn{4}{c}{195} & \multicolumn{4}{c}{195} & \multicolumn{4}{c}{195} \\
    $N_\textrm{selected}$ & \multicolumn{4}{c}{327} & \multicolumn{4}{c}{327} & \multicolumn{4}{c}{327} \\
    \bottomrule
    \end{tabular}
    \caption{Regression outcomes of Survey\_CoT on the music PWYW decision-making task. The table reports coefficients, standard errors, z-values, and p-values from a Heckman two-step model. Bottom panel presents model-level statistics including selection bias correction and sample distributions.}
    \label{tab:survey_cot_music}
\end{table*}

\paragraph{Survey Format – RAG}  
See Tables \ref{tab:survey_rag_art} and  \ref{tab:survey_rag_music}
\begin{table*}[]
    \centering
    \footnotesize
    \begin{tabular}{l|c@{\;\;}c@{\;\;}c@{\;\;}c|c@{\;\;}c@{\;\;}c@{\;\;}c|c@{\;\;}c@{\;\;}c@{\;\;}c}

    \toprule
    & \multicolumn{4}{c}{GPT-4o} & \multicolumn{4}{c}{LLaMA} & \multicolumn{4}{c}{Qwen} \\
         & coeff & std\_error& p value & z value & coeff & std\_error & p value &  z value  & coeff & std\_error & p value & z value \\
    \midrule
    \multicolumn{10}{l}{\textbf{Outcome}} \\
    edu\_f & 0.01 & 0.04 & 0.75 & -0.32      & 0.02 & 0.02 & 0.46 & 0.73      & 0.00 & 0.01 & 0.97 & -0.04\\
    edu\_m & -0.01 & 0.04 & 0.84 & -0.21      & 0.02 & 0.02 & 0.51 & 0.65      & 0.01 & 0.01 & 0.39 & 0.85 \\
    mus\_act\_dum & 0.08 & 0.09 & 0.36 & -1.59      & -0.03 & 0.05 & 0.56 & -0.58      & 0.00 & 0.02 & 0.88 & 0.15\\
    mus\_att\_ave & 0.13 & 0.08 & 0.11 & -0.37      & 0.06 & 0.05 & 0.21 & 1.26      & 0.00 & 0.02 & 0.81 & 0.25\\
    mus\_ca\_edu\_a19 & -0.01 & 0.03 & 0.71 & -0.57      & 0.00 & 0.02 & 0.98 & 0.03      & -0.01 & 0.01 & 0.08 & -1.73\\
    mus\_ca\_edu\_b18 & 0.02 & 0.04 & 0.57 & -0.17      & -0.02 & 0.02 & 0.32 & -0.99      & 0.00 & 0.01 & 0.92 & -0.10\\
    mus\_qz\_tot\_sco & -0.01 & 0.04 & 0.86 & 0.14      & 0.00 & 0.02 & 0.83 & -0.21      & -0.01 & 0.01 & 0.05 & -1.96\\
    moral\_n\_ave & 0.00 & 0.02 & 0.89 & -0.03      & -0.01 & 0.01 & 0.45 & -0.76      & 0.00 & 0.00 & 0.32 & -0.99\\
    social\_n\_ave & 0.00 & 0.02 & 0.97 & -0.06      & 0.00 & 0.01 & 0.81 & -0.24      & 0.01 & 0.00 & 0.17 & 1.36\\
    inc\_mo & 0.00 & 0.01 & 0.95 & 0.36      & 0.00 & 0.00 & 0.70 & -0.39      & 0.00 & 0.00 & 0.38 & 0.87\\
    matarial\_dum & 0.01 & 0.04 & 0.72 & 0.36      & 0.01 & 0.02 & 0.70 & -0.38      & 0.00 & 0.01 & 0.83 & -0.21\\
    \textit{constant} & 8.35 & 0.44 & p$<$0.0001 & 19.19      & 8.41 & 0.26 & p$<$0.0001 & -32.65      & 8.51 & 0.09 & p$<$0.0001& -95.45\\
    \midrule
    \multicolumn{10}{l}{\textbf{Model Statistics}} \\
    Wald $\chi^2$ & \multicolumn{4}{c}{10.45} & \multicolumn{4}{c}{28.81}& \multicolumn{4}{c}{15.87} \\
    $\rho$ & \multicolumn{4}{c}{1.00} & \multicolumn{4}{c}{-0.25}& \multicolumn{4}{c}{-0.33} \\
    $\sigma$ & \multicolumn{4}{c}{0.36} & \multicolumn{4}{c}{0.18}& \multicolumn{4}{c}{0.06} \\
    Log Likelihood & \multicolumn{4}{c}{-250.78} & \multicolumn{4}{c}{-250.78}& \multicolumn{4}{c}{-250.78} \\
    McFadden $R^2$ & \multicolumn{4}{c}{0.08} & \multicolumn{4}{c}{0.08}& \multicolumn{4}{c}{0.08} \\
    $N_\textrm{nonselected}$ & \multicolumn{4}{c}{113} & \multicolumn{4}{c}{113}& \multicolumn{4}{c}{113} \\
    $N_\textrm{selected}$ & \multicolumn{4}{c}{409} & \multicolumn{4}{c}{409}& \multicolumn{4}{c}{409} \\
    \bottomrule
    \end{tabular}
    \caption{Regression outcomes of Survey\_RAG on the art PWYW decision-making task. The table reports coefficients, standard errors, z-values, and p-values from a Heckman two-step model. Bottom panel presents model-level statistics including selection bias correction and sample distributions.}
    \label{tab:survey_rag_art}
\end{table*}

\begin{table*}[]
    \centering
    \footnotesize
    \begin{tabular}{l|c@{\;\;}c@{\;\;}c@{\;\;}c|c@{\;\;}c@{\;\;}c@{\;\;}c|c@{\;\;}c@{\;\;}c@{\;\;}c}

    \toprule
    & \multicolumn{4}{c}{GPT-4o} & \multicolumn{4}{c}{LLaMA} & \multicolumn{4}{c}{Qwen} \\
        & coeff & std\_error& p value & z value & coeff & std\_error & p value &  z value  & coeff & std\_error & p value & z value \\
    \midrule
    \multicolumn{10}{l}{\textbf{Outcome}} \\
    edu\_f & -0.05 & 0.03 & 0.14 & -1.49      & -0.06 & 0.03 & 0.06 & -1.85      & -0.08 & 0.06 & 0.14 & -1.48\\
    edu\_m & 0.00 & 0.03 & 0.97 & -0.03      & 0.02 & 0.03 & 0.56 & 0.58      & -0.01 & 0.05 & 0.86 & -0.18 \\
    mus\_act\_dum & -0.03 & 0.07 & 0.70 & -0.38      & -0.06 & 0.07 & 0.39 & -0.85      & -0.24 & 0.12 & p$<$0.05 & -2.02\\
    mus\_att\_ave & 0.02 & 0.10 & 0.82 & 0.23      & -0.05 & 0.10 & 0.61 & -0.51      & 0.11 & 0.17 & 0.53 & -0.63\\
    mus\_ca\_edu\_a19 & 0.03 & 0.03 & 0.23 & 1.21      & 0.02 & 0.03 & 0.58 & 0.55      & -0.05 & 0.05 & 0.27 & -1.11\\
    mus\_ca\_edu\_b18 & -0.01 & 0.03 & 0.71 & -0.38      & -0.04 & 0.03 & 0.14 & -1.49      & 0.00 & 0.05 & 0.95 & -0.06\\
    mus\_qz\_tot\_sco & 0.06 & 0.03 & p$<$0.05 & 2.05      & 0.03 & 0.03 & 0.30 & 1.04      & 0.00 & 0.06 & 0.94 & -0.07\\
    moral\_n\_ave & 0.04 & 0.02 & 0.13 & 1.51      & 0.03 & 0.02 & 0.26 & 1.12      & -0.07 & 0.05 & 0.13 & -1.51\\
    social\_n\_ave & 0.03 & 0.02 & 0.22 & 1.22      & 0.02 & 0.02 & 0.31 & 1.02      & 0.07 & 0.04 & 0.09 & -1.68\\
    inc\_mo & 0.01 & 0.01 & 0.14 & 1.49      & 0.00 & 0.01 & 0.65 & 0.46      & 0.01 & 0.02 & 0.38 & -0.88\\
    matarial\_dum & 0.01 & 0.03 & 0.82 & 0.23      & 0.01 & 0.04 & 0.88 & 0.15      & -0.06 & 0.06 & 0.35 & -0.93\\
    \textit{constant} & 9.32 & 0.41 & p$<$0.0001 & 22.70      & 10.08 & 0.41 & p$<$0.0001 & 24.34      & 9.81 & 0.73 & p$<$0.0001 & -13.47\\
    \midrule
    \multicolumn{10}{l}{\textbf{Model Statistics}} \\
    Wald $\chi^2$ & \multicolumn{4}{c}{30.46} & \multicolumn{4}{c}{14.85}& \multicolumn{4}{c}{22.56} \\
    $\rho$ & \multicolumn{4}{c}{-0.45} & \multicolumn{4}{c}{-1.00}& \multicolumn{4}{c}{-0.11} \\
    $\sigma$ & \multicolumn{4}{c}{0.30} & \multicolumn{4}{c}{0.35}& \multicolumn{4}{c}{0.51} \\
    Log Likelihood & \multicolumn{4}{c}{-290.01} & \multicolumn{4}{c}{-290.01}& \multicolumn{4}{c}{-590.01} \\
    McFadden $R^2$ & \multicolumn{4}{c}{0.16} & \multicolumn{4}{c}{0.16}& \multicolumn{4}{c}{0.16} \\
    $N_\textrm{nonselected}$ & \multicolumn{4}{c}{195} & \multicolumn{4}{c}{195}& \multicolumn{4}{c}{195} \\
    $N_\textrm{selected}$ & \multicolumn{4}{c}{327} & \multicolumn{4}{c}{327}& \multicolumn{4}{c}{327} \\
    \bottomrule
    \end{tabular}
    \caption{Regression outcomes of Survey\_RAG on the music PWYW decision-making task. The table reports coefficients, standard errors, z-values, and p-values from a Heckman two-step model. Bottom panel presents model-level statistics including selection bias correction and sample distributions.}
    \label{tab:survey_rag_music}
\end{table*}

\paragraph{Survey Format – Few-shot}  
See Tables \ref{tab:survey_fewshot_art} and \ref{tab:survey_cot_music} 
\begin{table*}[]
    \centering
    \footnotesize
    \begin{tabular}{l|c@{\;\;}c@{\;\;}c@{\;\;}c|c@{\;\;}c@{\;\;}c@{\;\;}c|c@{\;\;}c@{\;\;}c@{\;\;}c}

    \toprule
    & \multicolumn{4}{c}{GPT-4o} & \multicolumn{4}{c}{LLaMA} & \multicolumn{4}{c}{Qwen} \\
        & coeff & std\_error& p value & z value & coeff & std\_error & p value &  z value  & coeff & std\_error & p value & z value \\
    \midrule
    \multicolumn{10}{l}{\textbf{Outcome}} \\
    edu\_f & -0.22 & 0.34 & 0.53 & -0.63      & -0.04 & 0.40 & 0.92 & -0.10      & 0.03 & 0.03 & 0.46 & 0.74\\
    edu\_m & 0.26 & 0.40 & 0.51 & -0.66      & 0.05 & 0.46 & 0.91 & 0.11      & -0.03 & 0.04 & 0.51 & -0.65 \\
    mus\_act\_dum & -0.48 & 0.82 & 0.56 & -0.59      & 0.32 & 0.98 & 0.75 & 0.32      & 0.04 & 0.08 & 0.62 & -0.50\\
    mus\_att\_ave & -0.10 & 0.75 & 0.89 & -0.14      & -0.32 & 0.89 & 0.72 & -0.35      & 0.09 & 0.08 & 0.24 & 1.17\\
    mus\_ca\_edu\_a19 & -0.02 & 0.27 & 0.95 & -0.06      & -0.29 & 0.30 & 0.33 & -0.97      & 0.00 & 0.03 & 0.97 & -0.04\\
    mus\_ca\_edu\_b18 & -0.21 & 0.36 & 0.55 & -0.60      & -0.12 & 0.42 & 0.78 & -0.28      & 0.02 & 0.04 & 0.53 & 0.63\\
    mus\_qz\_tot\_sco & 0.12 & 0.34 & 0.73 & 0.34      & 0.16 & 0.40 & 0.68 & 0.41      & 0.02 & 0.03 & 0.55 & 0.59\\
    moral\_n\_ave & 0.21 & 0.20 & 0.31 & 1.02      & -0.25 & 0.25 & 0.32 & -0.99      & 0.00 & 0.02 & 0.93 & -0.08\\
    social\_n\_ave & -0.05 & 0.19 & 0.79 & -0.26      & 0.23 & 0.23 & 0.32 & 0.99      & 0.02 & 0.02 & 0.36 & 0.92\\
    inc\_mo & 0.03 & 0.08 & 0.70 & 0.38      & -0.09 & 0.09 & 0.35 & -0.94      & 0.00 & 0.01 & 0.57 & 0.56\\
    matarial\_dum & -0.03 & 0.32 & 0.92 & -0.09      & -0.75 & 0.36 & p$<$0.05 & -2.07      & 0.00 & 0.03 & 0.95 & 0.06\\
    \textit{constant} & 9.01 & 3.93 & p$<$0.05 & 2.30      & -8.45 & 4.71 & 0.07 & 1.79      & 8.44 & 0.40 & p$<$0.0001 & 21.17\\
    \midrule
    \multicolumn{10}{l}{\textbf{Model Statistics}} \\
    Wald $\chi^2$ & \multicolumn{4}{c}{3.82} & \multicolumn{4}{c}{21.44}& \multicolumn{4}{c}{5.55} \\
    rho & \multicolumn{4}{c}{-1.00} & \multicolumn{4}{c}{-0.76}& \multicolumn{4}{c}{1.00} \\
    $\sigma$ & \multicolumn{4}{c}{3.23} & \multicolumn{4}{c}{3.53}& \multicolumn{4}{c}{0.33} \\
    Log Likelihood & \multicolumn{4}{c}{-250.78} & \multicolumn{4}{c}{-250.78}& \multicolumn{4}{c}{-250.78} \\
    McFadden $R^2$ & \multicolumn{4}{c}{0.08} & \multicolumn{4}{c}{0.08}& \multicolumn{4}{c}{0.08} \\
    $N_\textrm{nonselected}$ & \multicolumn{4}{c}{113} & \multicolumn{4}{c}{113}& \multicolumn{4}{c}{113} \\
    $N_\textrm{selected}$ & \multicolumn{4}{c}{409} & \multicolumn{4}{c}{409}& \multicolumn{4}{c}{409} \\
    \bottomrule
    \end{tabular}
    \caption{Regression outcomes of Survey\_few-shot on the art PWYW decision-making task. The table reports coefficients, standard errors, z-values, and p-values from a Heckman two-step model. Bottom panel presents model-level statistics including selection bias correction and sample distributions.}
    \label{tab:survey_fewshot_art}
\end{table*}

\begin{table*}[]
    \centering
    \footnotesize
    \begin{tabular}{l|c@{\;\;}c@{\;\;}c@{\;\;}c|c@{\;\;}c@{\;\;}c@{\;\;}c|c@{\;\;}c@{\;\;}c@{\;\;}c}

    \toprule
    & \multicolumn{4}{c}{GPT-4o} & \multicolumn{4}{c}{LLaMA} & \multicolumn{4}{c}{Qwen} \\
         & coeff & std\_error& p value & z value & coeff & std\_error & p value &  z value  & coeff & std\_error & p value & z value \\
    \midrule
    \multicolumn{10}{l}{\textbf{Outcome}} \\
    edu\_f & -0.14 & 0.12 & 0.24 & -1.17      & -0.04 & 0.21 & 0.84 & -0.20      & -0.03 & 0.02 & 0.14 & -1.47\\
    edu\_m & 0.05 & 0.12 & 0.68 & -0.41      & -0.02 & 0.21 & 0.94 & -0.07      & 0.01 & 0.02 & 0.78 & 0.27 \\
    mus\_act\_dum & -0.21 & 0.28 & 0.45 & -0.75      & -0.36 & 0.47 & 0.45 & -0.75      & -0.01 & 0.05 & 0.90 & -0.13\\
    mus\_att\_ave & -0.27 & 0.36 & 0.46 & -0.74      & -0.54 & 0.62 & 0.39 & -0.87      & 0.06 & 0.07 & 0.44 & 0.77\\
    mus\_ca\_edu\_a19 & 0.01 & 0.12 & 0.93 & 0.09      & 0.01 & 0.20 & 0.98 & 0.03      & 0.04 & 0.02 & 0.08 & 1.74\\
    mus\_ca\_edu\_b18 & -0.05 & 0.11 & 0.67 & -0.43      & -0.10 & 0.18 & 0.59 & -0.54      & -0.02 & 0.02 & 0.42 & -0.80\\
    mus\_qz\_tot\_sco & -0.01 & 0.12 & 0.96 & -0.05      & 0.11 & 0.21 & 0.60 & 0.52      & 0.04 & 0.02 & 0.07 & 1.78\\
    moral\_n\_ave & 0.11 & 0.08 & 0.20 & 1.29      & 0.18 & 0.14 & 0.21 & 1.25      & 0.03 & 0.02 & 0.06 & 1.85\\
    social\_n\_ave & 0.01 & 0.08 & 0.91 & 0.11      & 0.08 & 0.14 & 0.57 & 0.57      & 0.01 & 0.02 & 0.45 & 0.76\\
    inc\_mo & 0.03 & 0.04 & 0.39 & 0.86      & 0.02 & 0.06 & 0.73 & -0.34      & 0.03 & 0.01 & p$<$0.0001 & 4.12\\
    matarial\_dum & 0.05 & 0.13 & 0.71 & 0.38      & -0.03 & 0.23 & 0.90 & -0.13      & -0.02 & 0.03 & 0.41 & -0.83\\
    \textit{constant} & 10.52 & 1.54 & p$<$0.0001 & 6.85      & 11.30 & 2.63 & p$<$0.0001 & 4.29      & 9.52 & 0.30 & p$<$0.0001 & -31.25\\
    \midrule
    \multicolumn{10}{l}{\textbf{Model Statistics}} \\
    Wald $\chi^2$ & \multicolumn{4}{c}{5.72} & \multicolumn{4}{c}{3.92}& \multicolumn{4}{c}{45.74} \\
    rho & \multicolumn{4}{c}{-1.00} & \multicolumn{4}{c}{-1.00}& \multicolumn{4}{c}{-0.55} \\
    $\sigma$ & \multicolumn{4}{c}{1.29} & \multicolumn{4}{c}{2.21}& \multicolumn{4}{c}{0.22} \\
    Log Likelihood & \multicolumn{4}{c}{-290.01} & \multicolumn{4}{c}{-290.01}& \multicolumn{4}{c}{-290.01} \\
    McFadden $R^2$ & \multicolumn{4}{c}{0.16} & \multicolumn{4}{c}{0.16}& \multicolumn{4}{c}{0.16} \\
    $N_\textrm{nonselected}$ & \multicolumn{4}{c}{195} & \multicolumn{4}{c}{195}& \multicolumn{4}{c}{195} \\
    $N_\textrm{selected}$ & \multicolumn{4}{c}{327} & \multicolumn{4}{c}{327}& \multicolumn{4}{c}{327} \\
    \bottomrule
    \end{tabular}
    \caption{Regression outcomes of Survey\_few-shot on the music PWYW decision-making task. The table reports coefficients, standard errors, z-values, and p-values from a Heckman two-step model. Bottom panel presents model-level statistics including selection bias correction and sample distributions.}
    \label{tab:survey_fewshot_music}
\end{table*}

\paragraph{Storytelling Format – Base}  
See Tables \ref{tab:story_base_art} and \ref{tab:story_base_music}
\begin{table*}[]
    \centering
    \footnotesize
    \begin{tabular}{l|c@{\;\;}c@{\;\;}c@{\;\;}c|c@{\;\;}c@{\;\;}c@{\;\;}c|c@{\;\;}c@{\;\;}c@{\;\;}c}

    \toprule
    & \multicolumn{4}{c}{GPT-4o} & \multicolumn{4}{c}{LLaMA} & \multicolumn{4}{c}{Qwen} \\
         & coeff & std\_error& p value & z value & coeff & std\_error & p value &  z value  & coeff & std\_error & p value & z value \\
    \midrule
    \multicolumn{10}{l}{\textbf{Outcome}} \\
    edu\_f & 0.02 & 0.07 & 0.76 & 0.30      & -0.07 & 0.08 & 0.38 & -0.87      & 0.00 & 0.00 & 0.73 & -0.35\\
    edu\_m & -0.01 & 0.08 & 0.91 & -0.11      & 0.09 & 0.09 & 0.30 & 1.03      & 0.00 & 0.00 & 0.20 & 1.27 \\
    mus\_act\_dum & 0.12 & 0.17 & 0.48 & 0.71      & -0.13 & 0.19 & 0.50 & -0.68      & 0.00 & 0.00 & 0.42 &-0.80 \\
    mus\_att\_ave & 0.19 & 0.15 & 0.21 & 1.25      & 0.04 & 0.17 & 0.82 & 0.22      & 0.00 & 0.00 & 0.36 & -0.92\\
    mus\_ca\_edu\_a19 & -0.03 & 0.05 & 0.53 & -0.63      & 0.01 & 0.06 & 0.82 & 0.22      & 0.00 & 0.00 & 0.13 & 1.53\\
    mus\_ca\_edu\_b18 & 0.05 & 0.07 & 0.50 & 0.68      & -0.03 & 0.08 & 0.74 & -0.33      & 0.00 & 0.00 & 0.75 & -0.32\\
    mus\_qz\_tot\_sco & 0.01 & 0.07 & 0.90 & 0.12      & -0.06 & 0.08 & 0.46 & -0.74      & 0.00 & 0.00 & 0.90 & -0.12\\
    moral\_n\_ave & 0.00 & 0.04 & 0.94 & -0.08      & -0.02 & 0.05 & 0.72 & -0.36      & 0.00 & 0.00 & 0.07 & 1.79\\
    social\_n\_ave & 0.00 & 0.04 & 0.91 & -0.11      & -0.01 & 0.04 & 0.83 & -0.21      & 0.00 & 0.00 & 0.63 & -0.48\\
    inc\_mo & -0.01 & 0.02 & 0.63 & -0.48      & 0.01 & 0.02 & 0.70 & 0.39      & 0.00 & 0.00 & 0.19 & -1.31\\
    matarial\_dum & 0.01 & 0.06 & 0.91 & 0.12      & -0.02 & 0.07 & 0.83 & -0.21      & 0.00 & 0.00 & 0.62 & -0.49\\
    \textit{constant} & 8.14 & 0.79 & p$<$0.0001 & 10.30      & 8.84 & 0.89 & p$<$0.0001 & 9.90      & 8.52 & 0.02 & p$<$0.0001 & 505.52\\
    \midrule
    \multicolumn{10}{l}{\textbf{Model Statistics}} \\
    Wald $\chi^2$ & \multicolumn{4}{c}{3.85} & \multicolumn{4}{c}{6.29}& \multicolumn{4}{c}{14.58} \\
    $\rho$ & \multicolumn{4}{c}{1.00} & \multicolumn{4}{c}{-1.00}& \multicolumn{4}{c}{-1.00} \\
    $\sigma$ & \multicolumn{4}{c}{0.65} & \multicolumn{4}{c}{0.74}& \multicolumn{4}{c}{0.01} \\
    Log Likelihood & \multicolumn{4}{c}{-250.78} & \multicolumn{4}{c}{-250.78}& \multicolumn{4}{c}{-250.78} \\
    McFadden $R^2$ & \multicolumn{4}{c}{0.08} & \multicolumn{4}{c}{0.08}& \multicolumn{4}{c}{0.08} \\
    $N_\textrm{nonselected}$ & \multicolumn{4}{c}{113} & \multicolumn{4}{c}{113}& \multicolumn{4}{c}{113} \\
    $N_\textrm{selected}$ & \multicolumn{4}{c}{409} & \multicolumn{4}{c}{409}& \multicolumn{4}{c}{409} \\
    \bottomrule
    \end{tabular}
    \caption{Regression outcomes of Storytelling\_Base on the art PWYW decision-making task. The table reports coefficients, standard errors, z-values, and p-values from a Heckman two-step model. Bottom panel presents model-level statistics including selection bias correction and sample distributions.}
    \label{tab:story_base_art}
\end{table*}

\begin{table*}[]
    \centering
    \footnotesize
    \begin{tabular}{l|c@{\;\;}c@{\;\;}c@{\;\;}c|c@{\;\;}c@{\;\;}c@{\;\;}c|c@{\;\;}c@{\;\;}c@{\;\;}c}

    \toprule
    & \multicolumn{4}{c}{GPT-4o} & \multicolumn{4}{c}{LLaMA} & \multicolumn{4}{c}{Qwen} \\
         & coeff & std\_error& p value & z value & coeff & std\_error & p value &  z value  & coeff & std\_error & p value & z value \\
    \midrule
    \multicolumn{10}{l}{\textbf{Outcome}} \\
    edu\_f & -0.03 & 0.03 & 0.32 & -0.99      & -0.03 & 0.05 & 0.53 & -0.63      & -0.07 & 0.03 & p$<$0.05 & -2.22\\
    edu\_m & -0.01 & 0.03 & 0.66 & -0.44      & -0.02 & 0.05 & 0.72 & -0.36      & -0.01 & 0.03 & 0.85 & -0.19 \\
    mus\_act\_dum & 0.00 & 0.06 & 0.97 & -0.03      & 0.02 & 0.12 & 0.87 & -0.17      & -0.06 & 0.07 & 0.43 & -0.79\\
    mus\_att\_ave & 0.11 & 0.09 & 0.20 & 1.27      & -0.11 & 0.16 & 0.52 & -0.64      & -0.04 & 0.10 & 0.66 & -0.44\\
    mus\_ca\_edu\_a19 & 0.04 & 0.03 & 0.08 & 1.74      & 0.04 & 0.05 & 0.45 & 0.75      & 0.01 & 0.03 & 0.66 & 0.44\\
    mus\_ca\_edu\_b18 & 0.00 & 0.02 & 0.95 & 0.06      & -0.03 & 0.05 & 0.49 & -0.69      & -0.02 & 0.03 & 0.48 & -0.71\\
    mus\_qz\_tot\_sco & 0.05 & 0.03 & 0.07 & 1.81      & 0.12 & 0.05 & p$<$0.05 & 2.32      & 0.04 & 0.03 & 0.22 & 1.22\\
    moral\_n\_ave & 0.02 & 0.02 & 0.30 & 1.05      & 0.13 & 0.04 & p$<$0.0001 & 3.01      & 0.05 & 0.02 & 0.05 & 2.00\\
    social\_n\_ave & 0.02 & 0.02 & 0.44 & 0.78      & -0.01 & 0.04 & 0.74 & -0.33      & 0.00 & 0.02 & 0.91 & 0.11\\
    inc\_mo & 0.01 & 0.01 & 0.26 & 1.13            & 0.02 & 0.02 & 0.23 & 1.21      & 0.01 & 0.01 & 0.34 & 0.96\\
    matarial\_dum & 0.01 & 0.03 & 0.77 & 0.30      & -0.06 & 0.06 & 0.32 & -1.00      & -0.03 & 0.03 & 0.34 & -0.95\\
    \textit{constant} & 9.07 & 0.37 & p$<$0.0001 & 24.47      & 9.77 & 0.70 & p$<$0.0001 & 14.05      & 9.91 & 0.40 & p$<$0.0001 & -24.67\\
    \midrule
    \multicolumn{10}{l}{\textbf{Model Statistics}} \\
    Wald $\chi^2$ & \multicolumn{4}{c}{31.27} & \multicolumn{4}{c}{31.22}& \multicolumn{4}{c}{26.14} \\
    $\rho$ & \multicolumn{4}{c}{0.41} & \multicolumn{4}{c}{-0.43}& \multicolumn{4}{c}{-0.87} \\
    $\sigma$ & \multicolumn{4}{c}{0.27} & \multicolumn{4}{c}{0.50}& \multicolumn{4}{c}{0.32} \\
    Log Likelihood & \multicolumn{4}{c}{-290.01} & \multicolumn{4}{c}{-290.01}& \multicolumn{4}{c}{-290.01} \\
    McFadden $R^2$ & \multicolumn{4}{c}{0.16} & \multicolumn{4}{c}{0.16}& \multicolumn{4}{c}{0.16} \\
    $N_\textrm{nonselected}$ & \multicolumn{4}{c}{195} & \multicolumn{4}{c}{195}& \multicolumn{4}{c}{195} \\
    $N_\textrm{selected}$ & \multicolumn{4}{c}{327} & \multicolumn{4}{c}{327}& \multicolumn{4}{c}{327} \\
    \bottomrule
    \end{tabular}
    \caption{Regression outcomes of Storytelling\_Base on the music PWYW decision-making task. The table reports coefficients, standard errors, z-values, and p-values from a Heckman two-step model. Bottom panel presents model-level statistics including selection bias correction and sample distributions.}
    \label{tab:story_base_music}
\end{table*}

\paragraph{Storytelling Format – CoT}  
See Tables \ref{tab:story_cot_art} and \ref{tab:story_cot_music}
\begin{table*}[]
    \centering
    \footnotesize
    \begin{tabular}{l|c@{\;\;}c@{\;\;}c@{\;\;}c|c@{\;\;}c@{\;\;}c@{\;\;}c|c@{\;\;}c@{\;\;}c@{\;\;}c}

    \toprule
    & \multicolumn{4}{c}{GPT-4o} & \multicolumn{4}{c}{LLaMA} & \multicolumn{4}{c}{Qwen} \\
         & coeff & std\_error& p value & z value & coeff & std\_error & p value &  z value  & coeff & std\_error & p value & z value \\
    \midrule
    \multicolumn{10}{l}{\textbf{Outcome}} \\
    edu\_f & -0.01 & 0.06 & 0.84 & -0.21      & 0.03 & 0.06 & 0.67 & 0.42      & 0.06 & 0.10 & 0.56 & 0.59\\
    edu\_m & 0.02 & 0.06 & 0.71 & 0.37        & 0.00 & 0.07 & 0.98 & -0.03      & -0.04 & 0.12 & 0.70 & -0.38 \\
    mus\_act\_dum & -0.07 & 0.13 & 0.58 & -0.55      & 0.15 & 0.15 & 0.31 & 1.01      & 0.20 & 0.24 & 0.41 & 0.82\\
    mus\_att\_ave & 0.02 & 0.12 & 0.86 & 0.18      & 0.14 & 0.14 & 0.29 & 1.05      & 0.21 & 0.22 & 0.34 & 0.95\\
    mus\_ca\_edu\_a19 & 0.00 & 0.04 & 0.96 & 0.05      & -0.01 & 0.05 & 0.82 & -0.23      & 0.03 & 0.08 & 0.71 & 0.37\\
    mus\_ca\_edu\_b18 & -0.03 & 0.06 & 0.57 & -0.57      & 0.05 & 0.07 & 0.45 & 0.75      & 0.09 & 0.10 & 0.37 & 0.90\\
    mus\_qz\_tot\_sco & -0.01 & 0.06 & 0.79 & -0.27      & 0.01 & 0.06 & 0.88 & 0.15      & 0.01 & 0.10 & 0.90 & 0.13\\
    moral\_n\_ave & 0.03 & 0.03 & 0.44 & 0.77      & 0.02 & 0.04 & 0.58 & 0.55          & 0.00 & 0.06 & 0.97 & -0.03\\
    social\_n\_ave & 0.00 & 0.03 & 0.97 & -0.03      & -0.01 & 0.03 & 0.71 & -0.37      & 0.00 & 0.05 & 0.96 & -0.05\\
    inc\_mo & 0.00 & 0.01 & 0.78 & 0.28              & 0.00 & 0.02 & 0.99 & 0.02      & 0.00 & 0.02 & 0.84 & 0.20\\
    matarial\_dum & -0.02 & 0.05 & 0.75 & -0.32      & 0.02 & 0.06 & 0.78 & 0.28      & 0.00 & 0.09 & 0.99 & -0.02\\
    \textit{constant} & 8.71 & 0.64 & p$<$0.0001 & 13.64      & 8.07 & 0.72 & p$<$0.0001 & 11.21      & 7.71 & 1.15 & p$<$0.0001 & 6.72\\
    \midrule
    \multicolumn{10}{l}{\textbf{Model Statistics}} \\
    Wald $\chi^2$ & \multicolumn{4}{c}{3.89} & \multicolumn{4}{c}{4.15}& \multicolumn{4}{c}{2.02} \\
    $\rho$ & \multicolumn{4}{c}{-1.00} & \multicolumn{4}{c}{1.00}& \multicolumn{4}{c}{1.00} \\
    $\sigma$ & \multicolumn{4}{c}{0.53} & \multicolumn{4}{c}{0.59}& \multicolumn{4}{c}{0.94} \\
    Log Likelihood & \multicolumn{4}{c}{-250.78} & \multicolumn{4}{c}{-250.78}& \multicolumn{4}{c}{-280.78} \\
    McFadden $R^2$ & \multicolumn{4}{c}{0.08} & \multicolumn{4}{c}{0.08}& \multicolumn{4}{c}{0.08} \\
    $N_\textrm{nonselected}$ & \multicolumn{4}{c}{113} & \multicolumn{4}{c}{113}& \multicolumn{4}{c}{113} \\
    $N_\textrm{selected}$ & \multicolumn{4}{c}{409} & \multicolumn{4}{c}{409}& \multicolumn{4}{c}{409} \\
    \bottomrule
    \end{tabular}
    \caption{Regression outcomes of Storytelling\_CoT on the art PWYW decision-making task. The table reports coefficients, standard errors, z-values, and p-values from a Heckman two-step model. Bottom panel presents model-level statistics including selection bias correction and sample distributions.}
    \label{tab:story_cot_art}
\end{table*}

\begin{table*}[]
    \centering
    \footnotesize
    \begin{tabular}{l|c@{\;\;}c@{\;\;}c@{\;\;}c|c@{\;\;}c@{\;\;}c@{\;\;}c|c@{\;\;}c@{\;\;}c@{\;\;}c}

    \toprule
    & \multicolumn{4}{c}{GPT-4o} & \multicolumn{4}{c}{LLaMA} & \multicolumn{4}{c}{Qwen} \\
         & coeff & std\_error& p value & z value & coeff & std\_error & p value &  z value  & coeff & std\_error & p value & z value \\
    \midrule
    \multicolumn{10}{l}{\textbf{Outcome}} \\
    edu\_f & -0.05 & 0.04 & 0.21 & -1.26          & -0.06 & 0.06 & 0.30 & -1.03      & 0.31 & 0.16 & 0.06 & 1.91\\
    edu\_m & 0.03 & 0.04 & 0.46 & 0.74            & -0.03 & 0.05 & 0.61 & -0.50      & -0.46 & 0.16 & p$<$0.0001 & -2.86 \\
    mus\_act\_dum & -0.01 & 0.08 & 0.90 & -0.12      & -0.03 & 0.12 & 0.78 & -0.27      & 0.15 & 0.36 & 0.67 & -0.42\\
    mus\_att\_ave & 0.04 & 0.11 & 0.71 & 0.37         & -0.07 & 0.17 & 0.68 & -0.42      & 0.69 & 0.50 & 0.17 & -1.37\\
    mus\_ca\_edu\_a19 & 0.04 & 0.03 & 0.27 & 1.09      & 0.01 & 0.05 & 0.77 & 0.29          & 0.22 & 0.15 & 0.14 & -1.47\\
    mus\_ca\_edu\_b18 & 0.00 & 0.03 & 0.98 & -0.03      & -0.02 & 0.05 & 0.61 & -0.51       & 0.06 & 0.14 & 0.68 & -0.41\\
    mus\_qz\_tot\_sco & 0.07 & 0.04 & 0.05 & 1.93      & 0.12 & 0.06 & p$<$0.05 & 2.05      & -0.22 & 0.17 & 0.18 & -1.34\\
    moral\_n\_ave & 0.09 & 0.03 & p$<$0.0001 & 2.97     & 0.14 & 0.04 & p$<$0.0001 & 3.14      & 0.11 & 0.13 & 0.40 & 0.83\\
    social\_n\_ave & -0.02 & 0.03 & 0.46 & -0.73      & 0.01 & 0.04 & 0.83 & 0.21             & 0.16 & 0.12 & 0.18 & 1.33\\
    inc\_mo & 0.01 & 0.01 & 0.21 & 1.25               & 0.03 & 0.02 & 0.07 & 1.78             & -0.02 & 0.05 & 0.75 & -0.32\\
    matarial\_dum & -0.03 & 0.04 & 0.50 & -0.67       & 0.06 & 0.06 & 0.32 & -1.00            & -0.15 & 0.18 & 0.40 & -0.84\\
    \textit{constant} & 9.19 & 0.48 & p$<$0.0001 & -19.15        & 9.50 & 0.73 & p$<$0.0001 & 13.00      & 6.44 & 2.12 & p$<$0.0001 & 3.03\\
    \midrule
    \multicolumn{10}{l}{\textbf{Model Statistics}} \\
    Wald $\chi^2$ & \multicolumn{4}{c}{27.60} & \multicolumn{4}{c}{34.69}& \multicolumn{4}{c}{17.90} \\
    $\rho$ & \multicolumn{4}{c}{-0.36} & \multicolumn{4}{c}{-0.51}& \multicolumn{4}{c}{0.67} \\
    $\sigma$ & \multicolumn{4}{c}{0.34} & \multicolumn{4}{c}{-0.53}& \multicolumn{4}{c}{1.60} \\
    Log Likelihood & \multicolumn{4}{c}{-290.01} & \multicolumn{4}{c}{-290.01}& \multicolumn{4}{c}{-290.01} \\
    McFadden $R^2$ & \multicolumn{4}{c}{0.16} & \multicolumn{4}{c}{-16}& \multicolumn{4}{c}{0.16} \\
    $N_\textrm{nonselected}$ & \multicolumn{4}{c}{195} & \multicolumn{4}{c}{195}& \multicolumn{4}{c}{195} \\
    $N_\textrm{selected}$ & \multicolumn{4}{c}{327} & \multicolumn{4}{c}{327}& \multicolumn{4}{c}{327} \\
    \bottomrule
    \end{tabular}
    \caption{Regression outcomes of Storytelling\_CoT on the music PWYW decision-making task. The table reports coefficients, standard errors, z-values, and p-values from a Heckman two-step model. Bottom panel presents model-level statistics including selection bias correction and sample distributions.}
    \label{tab:story_cot_music}
\end{table*}

\paragraph{Storytelling Format – RAG}  
See Tables \ref{tab:story_rag_art} and \ref{tab:story_rag_music}
\begin{table*}[]
    \centering
    \footnotesize
    \begin{tabular}{l|c@{\;\;}c@{\;\;}c@{\;\;}c|c@{\;\;}c@{\;\;}c@{\;\;}c|c@{\;\;}c@{\;\;}c@{\;\;}c}

    \toprule
    & \multicolumn{4}{c}{GPT-4o} & \multicolumn{4}{c}{LLaMA} & \multicolumn{4}{c}{Qwen} \\
         & coeff & std\_error& p value & z value & coeff & std\_error & p value &  z value  & coeff & std\_error & p value & z value \\
    \midrule
    \multicolumn{10}{l}{\textbf{Outcome}} \\
    edu\_f & 0.01 & 0.04 & 0.85 & 0.85             & 0.00 & 0.01 & 0.48 & 0.71      & - & - & - & -\\
    edu\_m & -0.01 & 0.04 & 0.85 & 0.85            & -0.01 & 0.01 & 0.29 & -1.06      & - & - & - & - \\
    mus\_act\_dum & 0.05 & 0.09 & 0.59 & 0.53      & 0.01 & 0.01 & 0.57 & 0.05      & - & - & - & -\\
    mus\_att\_ave & 0.12 & 0.08 & 0.15 & 1.44      & 0.01 & 0.01  & 0.36 & 0.92      & - & - & - & -\\
    mus\_ca\_edu\_a19 & -0.03 & 0.03 & 0.29 & -1.06      & 0.00 & 0.00 & 0.89 & 0.14       & - & - & - & -\\
    mus\_ca\_edu\_b18 & -0.01 & 0.04 & 0.89 & -0.14      & 0.00 & 0.01 & 0.66 & 0.45       & - & - & - & -\\
    mus\_qz\_tot\_sco & 0.01 & 0.04 & 0.72 & 0.36        & 0.01 & 0.01 & 0.39 & 0.87      & - & - & - & -\\
    moral\_n\_ave & 0.04 & 0.02 & 0.11 & 1.61            & 0.00 & 0.00 & 0.30 & 1.05       & - & - & - & -\\
    social\_n\_ave & -0.03 & 0.02 & 0.09 & -1.69         & 0.00 & 0.00 & 0.67 & -0.43       & - & - & - & -\\
    inc\_mo & 0.01 & 0.01 & 0.38 & 0.88              & 0.00 & 0.00 & 0.86 & 0.18      & - & - & - & -\\
    matarial\_dum & -0.01 & 0.03 & 0.72 & -0.36      & 0.00 & 0.01 & 0.81 & 0.24       & - & - & - & -\\
    \textit{constant} & 8.31 & 0.42 & p$<$0.0001 & 19.63        & 8.44 & 0.07 & p$<$0.0001 & 118.37       & - & - & - & -\\
    \midrule
    \multicolumn{10}{l}{\textbf{Model Statistics}} \\
    Wald $\chi^2$ & \multicolumn{4}{c}{21.11} & \multicolumn{4}{c}{5.10}& \multicolumn{4}{c}{-} \\
    $\rho$ & \multicolumn{4}{c}{0.72} & \multicolumn{4}{c}{0.99}& \multicolumn{4}{c}{-} \\
    $\sigma$ & \multicolumn{4}{c}{0.31} & \multicolumn{4}{c}{0.06}& \multicolumn{4}{c}{-} \\
    Log Likelihood & \multicolumn{4}{c}{-250.78} & \multicolumn{4}{c}{-250.78}& \multicolumn{4}{c}{-} \\
    McFadden $R^2$ & \multicolumn{4}{c}{0.08} & \multicolumn{4}{c}{0.08}& \multicolumn{4}{c}{-} \\
    $N_\textrm{nonselected}$ & \multicolumn{4}{c}{113} & \multicolumn{4}{c}{113}& \multicolumn{4}{c}{-} \\
    $N_\textrm{selected}$ & \multicolumn{4}{c}{409} & \multicolumn{4}{c}{409}& \multicolumn{4}{c}{-} \\
    \bottomrule
    \end{tabular}
    \caption{Regression outcomes of Storytelling\_RAG on the art PWYW decision-making task. The table reports coefficients, standard errors, z-values, and p-values from a Heckman two-step model. Bottom panel presents model-level statistics including selection bias correction and sample distributions. In Qwen, as all values were measured identitically, statistical analysis becomes impossible.}
    \label{tab:story_rag_art}
\end{table*}

\begin{table*}[]
    \centering
    \footnotesize
    \begin{tabular}{l|c@{\;\;}c@{\;\;}c@{\;\;}c|c@{\;\;}c@{\;\;}c@{\;\;}c|c@{\;\;}c@{\;\;}c@{\;\;}c}

    \toprule
    & \multicolumn{4}{c}{GPT-4o} & \multicolumn{4}{c}{LLaMA} & \multicolumn{4}{c}{Qwen} \\
         & coeff & std\_error& p value & z value & coeff & std\_error & p value &  z value  & coeff & std\_error & p value & z value \\
    \midrule
    \multicolumn{10}{l}{\textbf{Outcome}} \\
    edu\_f & -0.04 & 0.03 & 0.15 & -1.44                & -0.06 & 0.03 & p$<$0.05 & -2.36     & -0.03 & 0.04 & 0.43 & -0.79\\
    edu\_m & -0.01 & 0.03 & 0.83 & -0.21                & 0.02 & 0.02 & 0.52 & 0.65           & -0.04 & 0.04 & 0.26 & -1.12 \\
    mus\_act\_dum & -0.02 & 0.06 & 0.76 & -0.31         & -0.05 & 0.06 & 0.41 & -0.83         & -0.02 & 0.09 & 0.78 & -0.27\\
    mus\_att\_ave & 0.02 & 0.09 & 0.81 & 0.23           & -0.03 & 0.08 & 0.67 & -0.42         & 0.04 & 0.12 & 0.72 & 0.36\\
    mus\_ca\_edu\_a19 & 0.03 & 0.03 & 0.21 & 1.25          & 0.01 & 0.02 & 0.71 & 0.37        & 0.05 & 0.04 & 0.18 & 1.34\\
    mus\_ca\_edu\_b18 & -0.02 & 0.02 & 0.52 & -0.64        & -0.01 & 0.02 & 0.72 & -0.36      & -0.03 & 0.03 & 0.31 & -1.02\\
    mus\_qz\_tot\_sco & 0.05 & 0.03 & 0.10 & 1.64          & 0.05 & 0.03 & 0.06 & 1.90      & 0.04 & 0.04 & 0.35 & 0.94\\
    moral\_n\_ave & 0.03 & 0.02 & 0.13 & 1.44       & 0.04 & 0.02 & p$<$0.05 & 2.39         & 0.03 & 0.03 & 0.31 & 1.02\\
    social\_n\_ave & 0.02 & 0.01 & 0.22 & 0.77      & 0.01 & 0.02 & 0.65 & 0.45             & 0.02 & 0.03 & 0.41 & 0.98\\
    inc\_mo & 0.01 & 0.01 & 0.42 & 0.80             & 0.00 & 0.01 & 0.56 & 0.59             & 0.00 & 0.01 & 0.99 & 0.34\\
    matarial\_dum & 0.00 & 0.03 & 0.92 & 0.10       & 0.00 & 0.03 & 0.87 & 0.16             & 0.00 & 0.04 & p$<$0.05 & 0.01\\
    \textit{constant} & 9.46 & 0.37 & p$<$0.0001 & 25.32       & 9.89 & 0.33 & p$<$0.0001 & 30.24      & 9.57 & 0.52 & p$<$0.0001 & 18.48\\
    \midrule
    \multicolumn{10}{l}{\textbf{Model Statistics}} \\
    Wald $\chi^2$ & \multicolumn{4}{c}{23.42} & \multicolumn{4}{c}{26.57}& \multicolumn{4}{c}{22.24} \\
    $\rho$ & \multicolumn{4}{c}{-0.39} & \multicolumn{4}{c}{-0.82}& \multicolumn{4}{c}{-0.29} \\
    $\sigma$ & \multicolumn{4}{c}{0.27} & \multicolumn{4}{c}{0.26}& \multicolumn{4}{c}{0.37} \\
    Log Likelihood & \multicolumn{4}{c}{-290.01} & \multicolumn{4}{c}{-290.01}& \multicolumn{4}{c}{-290.01} \\
    McFadden $R^2$ & \multicolumn{4}{c}{0.16} & \multicolumn{4}{c}{0.16}& \multicolumn{4}{c}{0.16} \\
    $N_\textrm{nonselected}$ & \multicolumn{4}{c}{195} & \multicolumn{4}{c}{195}& \multicolumn{4}{c}{195} \\
    $N_\textrm{selected}$ & \multicolumn{4}{c}{327} & \multicolumn{4}{c}{327}& \multicolumn{4}{c}{327} \\
    \bottomrule
    \end{tabular}
    \caption{Regression outcomes of Storytelling\_RAG on the music PWYW decision-making task. The table reports coefficients, standard errors, z-values, and p-values from a Heckman two-step model. Bottom panel presents model-level statistics including selection bias correction and sample distributions.}
    \label{tab:story_rag_music}
\end{table*}

\section{RQ2 Heatmap}
\begin{figure*}[t]
    \centering
    \includegraphics[width=0.9\textwidth]{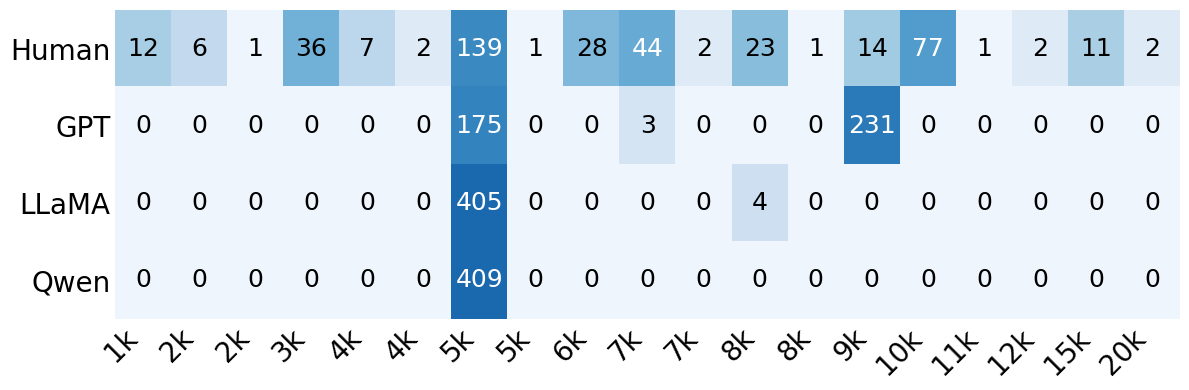}
    \caption{Response distribution in storytelling format $\times$ RAG condition}
    \label{fig:form6}
\end{figure*}

This section presents response distribution heatmaps for each experimental condition under RQ2. Figures from \ref{fig:form1} to \ref{fig:form6} visualize how LLM-generated willingness-to-pay responses vary depending on persona format (Survey vs. Storytelling) and prompting method (Base, CoT, RAG, Few-shot). Each figure corresponds to a specific combination (persona format, prompting method) and is provided to illustrate group-level tendencies and response clustering patterns across conditions.
\paragraph{Survey Format - CoT}
See Figure~\ref{fig:form1}
\begin{figure*}[t]
    \centering
    \includegraphics[width=0.9\textwidth]{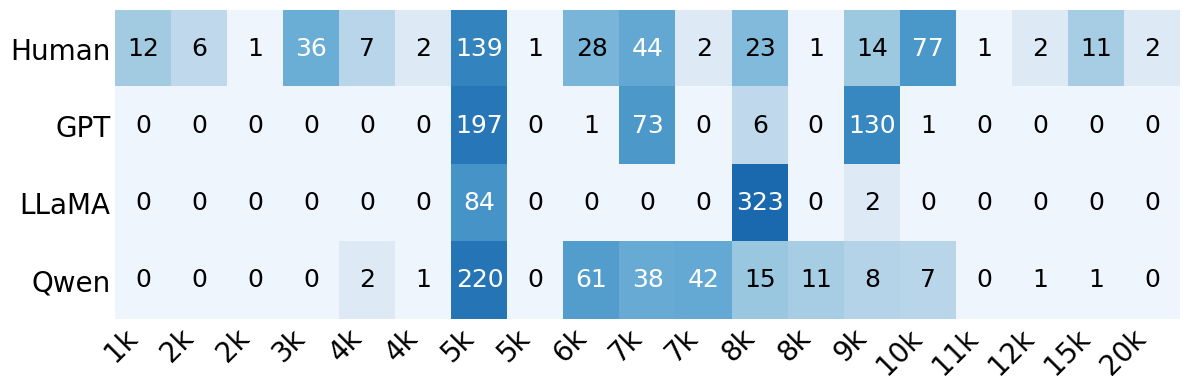}
    \caption{Response distribution in survey format $\times$ CoT condition}
    \label{fig:form1}
\end{figure*}

\paragraph{Survey Format - RAG}
See Figure~\ref{fig:form2}
\begin{figure*}[t]
    \centering
    \includegraphics[width=0.9\textwidth]{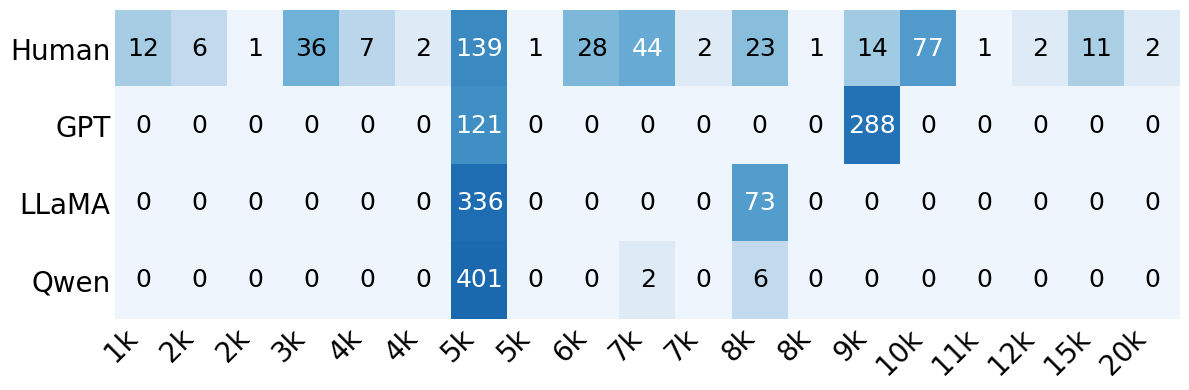}
    \caption{Response distribution in survey format $\times$ RAG condition}
    \label{fig:form2}
\end{figure*}

\begin{figure*}[t]
    \centering
    \includegraphics[width=0.9\textwidth]{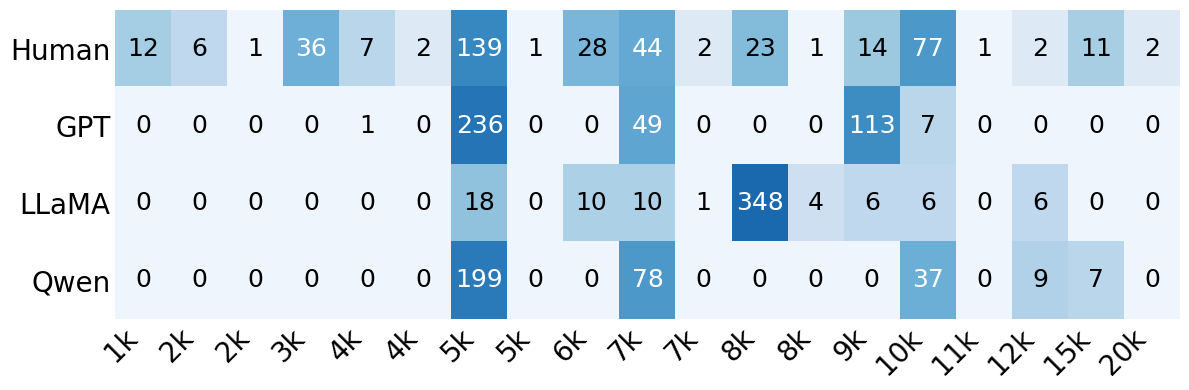}
    \caption{Response distribution in survey format $\times$ few-shot condition}
    \label{fig:form3}
\end{figure*}

\paragraph{Survey Format - Few Shot}
See Figure~\ref{fig:form3}

\begin{figure*}[t]
    \centering
    \includegraphics[width=0.9\textwidth]{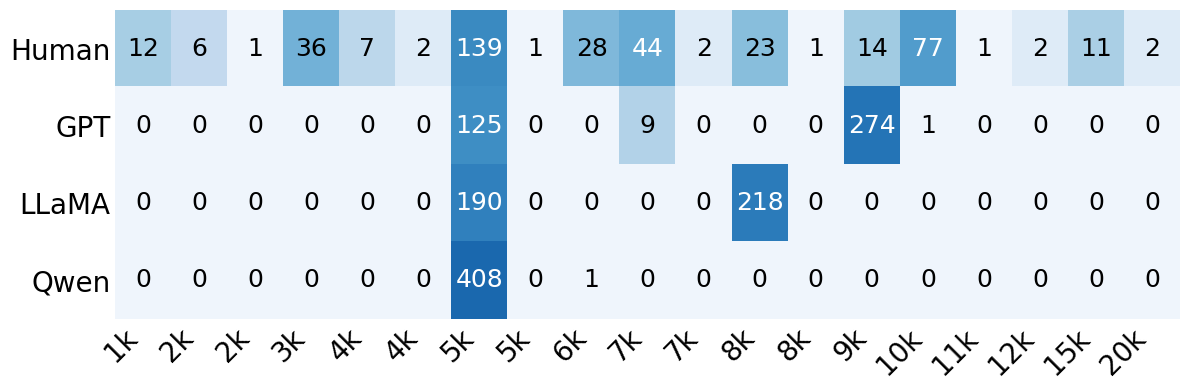}
    \caption{Response distribution in storytelling format $\times$ base condition}
    \label{fig:form4}
\end{figure*}

\paragraph{Storytelling Format - Base}
See Figure~\ref{fig:form4}

\begin{figure*}[t]
    \centering
    \includegraphics[width=0.9\textwidth]{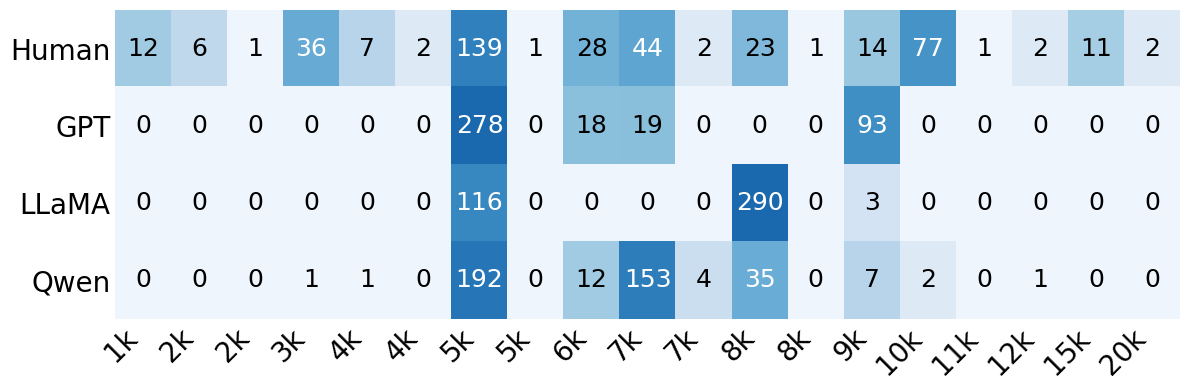}
    \caption{Response distribution in storytelling format $\times$ CoT condition}
    \label{fig:form5}
\end{figure*}

\paragraph{Storytelling Format - CoT}
See Figure~\ref{fig:form5}

\paragraph{Storytelling Format - RAG}
See Figure~\ref{fig:form6}

\end{document}